\documentclass[sigplan,10pt]{acmart}
\acmSubmissionID{}
\renewcommand\footnotetextcopyrightpermission[1]{}

\hypersetup{colorlinks,linkcolor=blue!60!black,citecolor=blue!60!black,urlcolor=blue!60!black}
\usepackage{xspace}
\usepackage{subcaption}
\usepackage{booktabs}
\usepackage{multirow}
\usepackage[linesnumbered,ruled,vlined,noend]{algorithm2e}
\usepackage{soul}
\usepackage{amsmath,amsfonts}
\usepackage{graphicx}
\usepackage{xcolor}
\usepackage{enumitem}
\usepackage{bm}
\usepackage{pifont}

\newcommand{\cmark}{\ding{51}}

\newcommand{\sys}[0]{{FairyFuse}\xspace}

\newcommand{\R}{\mathbb{R}}

\begin{document}

\settopmatter{printacmref=false}
\acmYear{2026}\copyrightyear{2026}
\setcopyright{none}
\acmConference[]{}{}{}
\acmBooktitle{}
\acmDOI{}
\acmISBN{}

\title{FairyFuse: Multiplication-Free LLM Inference on CPUs via Fused Ternary Kernels}

\settopmatter{authorsperrow=5}

\author{Fei Zuo}
\authornote{Equal contribution.}
\affiliation{%
  \institution{BA TechWorks}
  \department{BMW Group}
  \country{}
}

\author{Xiaoyan Xi}
\authornotemark[1]
\affiliation{%
  \institution{BA TechWorks}
  \department{BMW Group}
  \country{}
}

\author{Quanyi Zeng}
\authornotemark[1]
\affiliation{%
  \institution{Peking University}
  \country{}
}

\author{Feiyu Wang}
\affiliation{%
  \institution{Peking University}
  \country{}
}

\author{Ho Fai Leung}
\authornote{Corresponding author.}
\affiliation{%
  \institution{BA TechWorks}
  \department{BMW Group}
  \country{}
}

\begin{abstract}
Large language models are increasingly deployed on CPU-only platforms, from on-device assistants to privacy-sensitive edge servers, where memory bandwidth is the dominant bottleneck for autoregressive token generation.
Weight quantization to four bits or below reduces memory pressure, yet existing inference systems still dequantize the compressed weights and execute floating-point multiplications, leaving much of the potential bit-width savings unrealized.
Ternary weights, which take values in $\{-1,0,+1\}$, offer a more radical alternative: every weight-activation product can be replaced by a conditional addition, subtraction, or no-op, eliminating multiplications altogether.
Fairy2i recently demonstrated that complex-valued quantization-aware training can produce ternary LLMs whose quality rivals FP16, yet its reference runtime still dispatches the quantized weights through standard floating-point routines, leaving the ternary structure unexploited at the hardware level.
Here we present \sys{}, a companion inference system for Fairy2i that realizes the multiplication-free promise on commodity CPUs for the first time.
\sys{} fuses the eight real-valued sub-GEMVs of each widely-linear layer into a single AVX-512 loop driven entirely by masked additions and subtractions; assembly inspection confirms that the inner loop contains zero floating-point multiplication instructions.
Roofline analysis reveals a structural asymmetry: the $16\times$ data compression from ternary packing shifts deeply memory-bound GEMV operations toward the compute ridge on bandwidth-limited CPUs, yielding a $29.6\times$ kernel speedup, whereas the same compression provides negligible benefit on bandwidth-rich GPUs.
End-to-end, \sys{} sustains 32.4 tokens per second on a single Intel Xeon 8558P socket, outperforming llama.cpp Q4\_K\_M by $1.24\times$ while preserving near-lossless quality (WikiText-2 perplexity 5.52 versus 5.47 for FP16; average downstream accuracy 66.0\%).
All measurements are reproducible with a coefficient of variation below 2\%.
\end{abstract}

\maketitle



\pagestyle{empty}

\section{Introduction}

{\sloppy
Large language models (LLMs)~\citep{vaswani2017attention,touvron2023llama2} are increasingly deployed beyond datacenters: on-device assistants, desktop copilots, and embedded agents all demand low-latency, privacy-preserving inference without discrete GPUs.
On such platforms the CPU is the only available compute engine, and memory bandwidth is the dominant bottleneck for autoregressive decoding~\citep{pope2023efficiently}.
Weight quantization is the primary lever for fitting billion-parameter models into limited DRAM and for keeping the weight-streaming working set cache-feasible.
\par}

\begin{figure*}[t]
  \centering
  \includegraphics[width=0.95\linewidth]{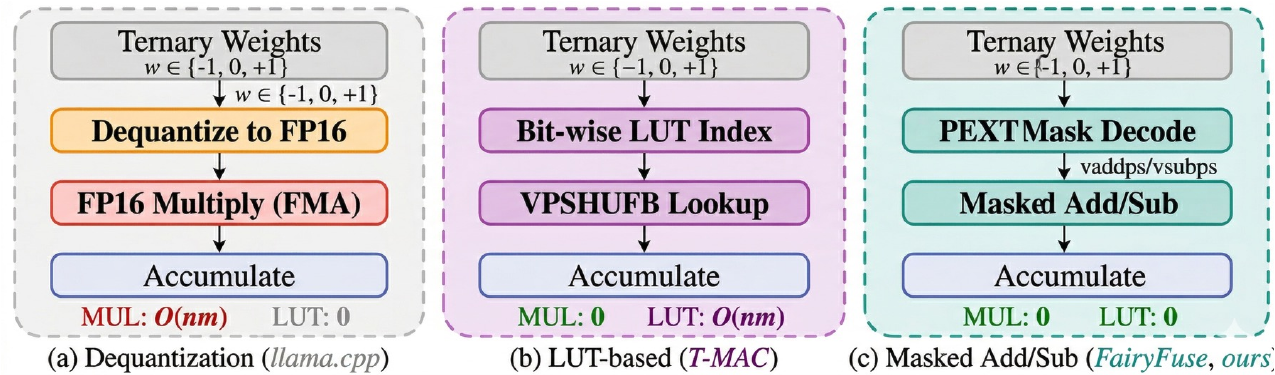}
  \vspace{-0.5em}
  \caption{\sys{} compared with existing approaches for ternary LLM inference.
  (a)~Conventional systems dequantize ternary weights to FP16 and perform standard multiplication.
  (b)~LUT-based systems (e.g., T-MAC) replace multiplications with table lookups but incur on-chip memory pressure.
  (c)~\sys{} directly applies masked addition and subtraction on packed 2-bit weights, with zero multiplications and zero lookup-table overhead.}
  \vspace{-1em}
  \label{fig:intro}
\end{figure*}

At the extreme end of this spectrum, ternary weights $\{-1,0,+1\}$ achieve $16\times$ compression over FP32, and each weight-activation product reduces to a conditional addition, subtraction, or no-op, eliminating multiplications entirely.
This idea has roots in early ternary~\citep{li2016ternary,zhu2017trained} and binary~\citep{rastegari2016xnor,hubara2016binarized} networks, and recent work confirms that ternary LLMs can match FP16 quality: BitNet~b1.58~\cite{ma2024era} trains models from scratch at ternary precision, while Fairy2i~\cite{feng2025fairy2i} post-trains existing checkpoints into complex-valued ternary models via quantization-aware training.
Fairy2i solves the algorithmic side of the problem; the systems challenge is to translate ternary weights into multiplication-free execution on real hardware.

Existing inference systems, however, do not yet exploit the ternary structure at the microarchitectural level.
Fairy2i's reference runtime, designed primarily as a research prototype for validating model quality, dispatches the quantized weights through standard PyTorch linear operators, leaving room for a dedicated inference engine that eliminates multiplications at the instruction level.
Similarly, llama.cpp~\cite{llamacpp2024} applies dequantization followed by floating-point FMA for its Q2\_K and Q4\_K\_M formats, so the low-bit encoding serves only as a storage optimization while the compute path remains multiply-heavy.
An alternative line of work replaces multiplication with bitwise lookup tables: T-MAC~\cite{wei2024tmac} and BitNet.cpp~\cite{bitnetcpp2024} use \texttt{VPSHUFB}-based LUT lookups, but the resulting kernels incur indirect memory accesses and non-trivial on-chip LUT footprints that compete with activation data for L1/L2 cache bandwidth, limiting sustained throughput on wide SIMD units.

Our starting observation is that ternary arithmetic on x86 CPUs admits a strictly simpler datapath than either dequantization or table lookup.
We pack sixteen $\{-1,0,+1\}$ weights into a single 32-bit word, decode positive and negative masks in a single-cycle \texttt{\_pext\_u32} pair (BMI2), and accumulate via AVX-512 masked \texttt{vaddps}/\texttt{vsubps}.
The resulting inner loop contains \emph{zero} floating-point multiplications and \emph{zero} table lookups, a property we verify through assembly inspection.

Realizing this idea for complex-valued models introduces an additional complication.
Fairy2i's widely-linear representation decomposes each linear layer into \emph{eight} coupled real-valued sub-GEMVs.
A naive schedule launches eight separate parallel regions, repeatedly loads the same activation tiles, spills intermediate buffers to DRAM, and re-decodes related mask patterns, erasing the bandwidth gains from ternary packing.
To overcome this overhead, we design a \emph{Fused Widely-Linear Kernel} that consolidates all eight sub-GEMVs into one row-parallel AVX-512 loop.
Four design decisions make fusion profitable: each decoded mask pair serves both real and imaginary accumulation paths, amortizing decode cost (O1); activation registers are loaded once and shared across all eight sub-GEMVs (O2); complex conjugation reduces to a precomputed sign-flipped alias, eliminating per-lane negation (O3); and all partial sums remain register-resident until the final horizontal reduction, removing intermediate DRAM traffic (O4).
Roofline analysis confirms that the $16\times$ compression shifts GEMV from deeply memory-bound (AI${}=0.25$) toward the CPU ridge point (AI${}=8$), yielding a $29.6\times$ speedup over FP32, while the same compression provides negligible benefit on bandwidth-rich GPUs.

End-to-end, \sys{} sustains 32.4~tok/s on a single Intel Xeon 8558P socket, $1.24\times$ faster than llama.cpp Q4\_K\_M, while preserving near-lossless quality: WikiText-2 perplexity is 5.52 (vs.\ 5.47 FP16) and average downstream accuracy is 66.0\%.
Assembly analysis confirms zero multiplication instructions in the inner loop, and all measurements are reproducible with CV${}<{}$2\%.

In summary, this paper makes three contributions:
\begin{itemize}
  \item We implement the first ternary-weight GEMV kernel on x86 CPUs that uses only masked AVX-512 additions and subtractions, with no floating-point multiplications or table lookups in the inner loop.
  \item We consolidate the eight sub-GEMVs of a complex-valued widely-linear layer into a single SIMD-friendly loop via mask reuse, input sharing, sign-swap aliasing, and register-resident accumulation, achieving $1.55\times$ speedup over unfused execution.
  \item Through full-stack evaluation on a 7B-parameter model, we demonstrate that ternary packing makes CPUs, not GPUs, the natural target for extreme quantization, and deliver throughput competitive with 4-bit baselines at 2-bit storage.
\end{itemize}

The rest of the paper is organized as follows.
Section~\ref{sec:background} provides background on CPU inference, ternary quantization, and the widely-linear GEMV structure.
Section~\ref{sec:design} presents the \sys{} algorithm and fused kernel design.
Section~\ref{sec:impl} describes the implementation.
Section~\ref{sec:eval} evaluates kernel performance, end-to-end throughput and quality, Roofline analysis, and design ablations.
Section~\ref{sec:related} discusses related work, and Section~7 concludes.

\section{Background and Motivation}
\label{sec:background}

\subsection{LLM Inference on CPUs}

Autoregressive large language model inference~\citep{vaswani2017attention} is dominated by the decode phase, where each new token is produced by applying a stack of linear layers to a short activation vector.
In this regime, the critical operation is a batched matrix-vector product (GEMV) whose cost is set almost entirely by how quickly model weights can be streamed from DRAM\@.
Concretely, generating one token requires touching every weight parameter at least once, so throughput scales with memory bandwidth rather than peak floating-point throughput~\citep{pope2023efficiently}.
On commodity CPUs with approximately 200~GB/s of DRAM bandwidth, this bottleneck is especially acute: even aggressively optimized FP16 or INT8 kernels remain far below arithmetic ceilings because the memory system cannot feed operands fast enough.

\begin{figure*}[t]
  \centering
  \includegraphics[width=0.95\linewidth]{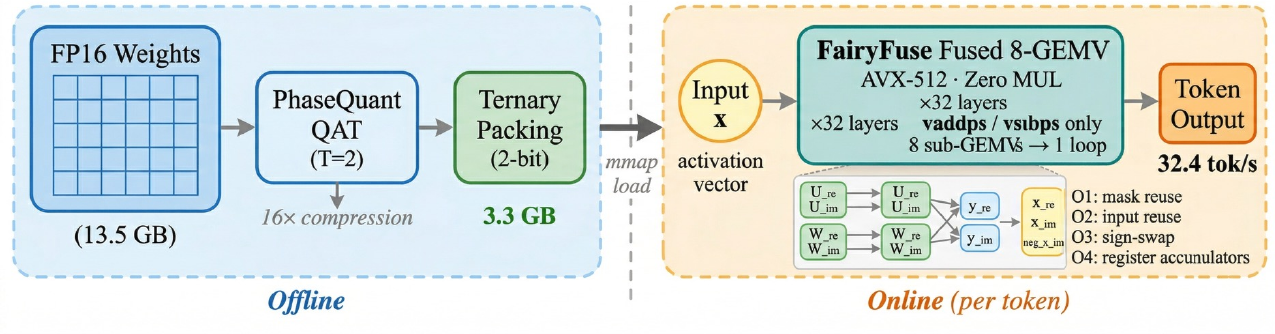}
  \vspace{-0.5em}
  \caption{System overview of \sys{}. Offline, Fairy2i's quantization converts FP16 weights to 2-bit packed ternary format (3.3\,GB).
  Online, the FairyFuse kernel performs fused 8-GEMV using only masked AVX-512 additions and subtractions, achieving 32.4\,tok/s.}
  \vspace{-1em}
  \label{fig:overview}
\end{figure*}

CPU-based inference nevertheless remains attractive for many deployment settings.
General-purpose processors are ubiquitous on laptops, desktops, and embedded systems; they avoid the power, thermal, and binary-stack constraints of discrete accelerators and simplify compliance with data sovereignty requirements by keeping models entirely on the host.
Low-latency, privacy-preserving, and fully offline assistants are therefore natural targets for CPU execution, provided that inference throughput can be pushed high enough to be interactive at billion-parameter scales.

\subsection{Ternary Quantization for LLMs}

Post-training and quantization-aware methods have substantially reduced the precision of weights and activations while preserving model quality.
Representative families include post-training quantization for generative transformers~\citep{frantar2023gptq,yao2022zeroquant}, activation-aware weight quantization~\citep{lin2024awq}, and accurate 8-bit schemes with careful scaling~\citep{xiao2023smoothquant}.
These approaches typically operate at four to eight bits per weight and still rely on multiply-accumulate hardware at inference time.

At the extreme, \emph{ternary} weights in $\{-1,0,+1\}$ occupy only 2~bits, a $16\times$ footprint reduction versus FP32, and replace each weight-activation product with a conditional add, subtract, or no-op.
BitNet~b1.58~\citep{ma2024era} showed that such ternary LLMs can match FP16 quality when trained appropriately.
Closest to our setting, Fairy2i~\citep{feng2025fairy2i,feng2024ifairy} encodes pretrained FP16 models into complex-valued ternary tensors through a lightweight quantization-aware training procedure called PhaseQuant, often requiring only a couple of calibration epochs.
Weights live in $\{1,-1,i,-i\}$ in the complex domain; equivalently, each real and imaginary component is ternary in $\{-1,0,+1\}$, yielding the same 2-bit storage budget while exploiting phase structure during training.
\sys{} is designed as the inference engine for these Fairy2i models, converting the algorithmic promise of ternary weights into actual multiplication-free execution.

\subsection{Complex-Valued Widely-Linear GEMV}

Complex-valued neural networks~\citep{trabelsi2018deep} have shown benefits in signal processing and representation learning.
A widely linear transformation augments a standard complex linear map with a conjugate term~\citep{picinbono1995widely}.
For complex weight matrices $\mathbf{U},\mathbf{W}\in\mathbb{C}^{n\times m}$ and activations $\mathbf{x}\in\mathbb{C}^{m}$, a widely linear layer computes
\begin{equation}
  \mathbf{y} = \mathbf{U}\mathbf{x} + \mathbf{W}\overline{\mathbf{x}}\,,
  \label{eq:widely_linear}
\end{equation}
where $\overline{\mathbf{x}}$ denotes complex conjugation.
Decomposing $\mathbf{x}=\mathbf{x}_{\mathrm{re}}+i\mathbf{x}_{\mathrm{im}}$ and writing $\mathbf{U}=\mathbf{U}_{\mathrm{re}}+i\mathbf{U}_{\mathrm{im}}$, $\mathbf{W}=\mathbf{W}_{\mathrm{re}}+i\mathbf{W}_{\mathrm{im}}$ with learned per-channel scales $s^{U}_{\mathrm{re}}$, $s^{U}_{\mathrm{im}}$, $s^{W}_{\mathrm{re}}$, $s^{W}_{\mathrm{im}}$, one can separate real and imaginary outputs into eight real-valued GEMVs:
\begin{align}
  \mathbf{y}_{\mathrm{re}}
  &= s^{U}_{\mathrm{re}}\,\mathbf{U}_{\mathrm{re}}\mathbf{x}_{\mathrm{re}}
   - s^{U}_{\mathrm{im}}\,\mathbf{U}_{\mathrm{im}}\mathbf{x}_{\mathrm{im}}
   + s^{W}_{\mathrm{re}}\,\mathbf{W}_{\mathrm{re}}\mathbf{x}_{\mathrm{re}}
   + s^{W}_{\mathrm{im}}\,\mathbf{W}_{\mathrm{im}}\mathbf{x}_{\mathrm{im}}\,,
  \label{eq:y_re} \\
  \mathbf{y}_{\mathrm{im}}
  &= s^{U}_{\mathrm{re}}\,\mathbf{U}_{\mathrm{re}}\mathbf{x}_{\mathrm{im}}
   + s^{U}_{\mathrm{im}}\,\mathbf{U}_{\mathrm{im}}\mathbf{x}_{\mathrm{re}}
   + s^{W}_{\mathrm{re}}\,\mathbf{W}_{\mathrm{re}}\mathbf{x}_{\mathrm{im}}
   - s^{W}_{\mathrm{im}}\,\mathbf{W}_{\mathrm{im}}\mathbf{x}_{\mathrm{re}}\,.
  \label{eq:y_im}
\end{align}

Because every weight block is ternary, each of the eight GEMVs reduces to masked additions and subtractions over activations: the $O(nm)$ dominant work is \emph{multiplication-free}.
Only the $O(n)$ application of scalar scales to partial sums retains floating-point multiplies, which is asymptotically negligible relative to the inner dimension for large language models.

\subsection{Deployment Challenges of Ternary LLMs}

Despite the arithmetic structure above, practical systems have not yet translated it into multiplication-free execution on real hardware.
Fairy2i's reference implementation focuses on validating model quality and naturally relies on standard PyTorch operators (\texttt{F.linear}), which dispatch the quantized weights through conventional multiply-accumulate routines; the ternary structure remains unexploited at the instruction level.
Alternative low-bit CPU engines such as BitNet.cpp and T-MAC replace multiplications with table-driven decoding~\citep{wei2024tmac}, but LUT kernels trade arithmetic for irregular table lookups that increase L1 pressure and can limit sustained bandwidth on wide SIMD units.

Even when ternary structure is respected, the widely-linear form in Eqs.~\eqref{eq:widely_linear}--\eqref{eq:y_im} introduces a second challenge: each complex layer maps to eight real-valued GEMVs.
A naive implementation issues eight separate parallel regions (or kernel launches), multiplying synchronization, loop overhead, and register-spill pressure, and may repeatedly load the same activation tiles from memory for different sub-products.
These overheads erode the theoretical gains from ternary arithmetic and complicate integration with existing LLM inference frameworks.
The design of \sys{} directly targets both issues: multiplication-free ternary GEMV on real CPUs and fusion across the eight sub-GEMVs, so that complex-valued quantization becomes a deployment asset rather than a constant-factor slowdown.

\section{Design}
\label{sec:design}

Figure~\ref{fig:overview} summarizes the \sys{} system.
The offline stage extracts ternary weights from the Fairy2i-W2 checkpoint and packs them into the 2-bit encoding described below, yielding a 3.3\,GB footprint for seven billion parameters ($16\times$ smaller than FP32).
During online decoding, each generated token passes through 32 Transformer layers; every linear projection that implements the widely-linear map of Eqs.~\eqref{eq:widely_linear}--\eqref{eq:y_im} is evaluated by the \sys{} fused AVX-512 kernel, which performs all eight real-valued GEMVs as a single row-parallel loop (Section~\ref{sec:fused}).
We describe the ternary GEMV algorithm first, then present the fused kernel design.

\subsection{\sys{} Algorithm}
\label{sec:sys-alg}

Let $w\in\{-1,0,+1\}$ be a ternary weight and $a\in\R$ an activation.
The product decomposes without any multiply as
$w\cdot a = a$ if $w{=}+1$, $-a$ if $w{=}-1$, and $0$ if $w{=}0$.
We pack sixteen weights into one \texttt{uint32} using two bits $(b_1,b_0)$
per weight: $(1,0)$ encodes $+1$ (\emph{add}), $(0,1)$ encodes $-1$
(\emph{subtract}), and $(0,0)$ encodes $0$; the pattern $(1,1)$ is unused.
This representation achieves $16\times$ compression relative to FP32 while
keeping decode logic SIMD-friendly.

\paragraph{Parallel bit decode.}
Let $p\in\{0,\ldots,2^{32}{-}1\}$ denote one packed word.
Two selection masks pick alternating bit lanes in
BMI2 \texttt{\_pext\_u32} (parallel bits extract):
\begin{equation*}
  M_{\mathrm{pos}} = \texttt{0xAAAAAAAAu},\qquad
  M_{\mathrm{neg}} = \texttt{0x55555555u}.
\end{equation*}
\vspace{-0.5em}
We obtain 16-bit masks over one 512-bit SIMD lane by
\begin{align}
  k_{\mathrm{pos}} &= \texttt{\_pext\_u32}(p,\, M_{\mathrm{pos}})\,,
  \label{eq:pext_pos}\\
  k_{\mathrm{neg}} &= \texttt{\_pext\_u32}(p,\, M_{\mathrm{neg}})\,.
  \label{eq:pext_neg}
\end{align}
Intuitively, $k_{\mathrm{pos}}$ collects the ``add'' bits and
$k_{\mathrm{neg}}$ the ``subtract'' bits after compaction into contiguous
sub-registers.

\paragraph{Masked accumulation.}
Given a 16-wide FP32 chunk $\mathbf{x}\in\R^{16}$ loaded into an AVX-512
register $X$, the ternary multiply-accumulate reduces to two masked floating-point operations:
\begin{align}
  \mathbf{acc} &\leftarrow
  \texttt{\_mm512\_mask\_add\_ps}(\mathbf{acc},\, k_{\mathrm{pos}},\,
  \mathbf{acc},\, X)\,,
  \label{eq:mask_add}\\
  \mathbf{acc} &\leftarrow
  \texttt{\_mm512\_mask\_sub\_ps}(\mathbf{acc},\, k_{\mathrm{neg}},\,
  \mathbf{acc},\, X)\,.
  \label{eq:mask_sub}
\end{align}
Each 16-element chunk therefore executes \emph{zero} floating-point
multiplications: only bitwise extract, mask logic, and masked add/sub
operations touch the execution units.

\paragraph{Comparison to T-MAC.}
T-MAC~\citep{wei2024tmac} rewrites low-bit convolutions into table
lookups and feeds lookup results through \texttt{VPSHUFB}, retaining an
indirect memory access and a non-zero LUT footprint in cache.
Our scheme has \emph{no} LUT: masks drive SIMD units directly, avoiding
gather/scatter-style shuffles on activation-dependent tables and
eliminating the associated TLB and cache pressure.
Where T-MAC's bit-wise LUT path issues packed byte shuffles, we issue a
single-cycle \texttt{\_pext} pair followed by AVX-512 masked add/sub on
the live activation vector, a structurally different datapath aligned with
ternary $\{-1,0,+1\}$ arithmetic rather than small-integer product tables.

\SetKwInOut{KwIn}{Input}
\SetKwInOut{KwOut}{Output}
\begin{algorithm}[t]
\caption{\sys{} Fused 8-GEMV (one output row $i$)}
\label{alg:sys}
\small
\DontPrintSemicolon
\KwIn{Packed rows $(\mathbf{U}_{\mathrm{re}})_{i,\cdot},(\mathbf{U}_{\mathrm{im}})_{i,\cdot},
(\mathbf{W}_{\mathrm{re}})_{i,\cdot},(\mathbf{W}_{\mathrm{im}})_{i,\cdot} \in
\texttt{uint32}^{m/16}$;
vectors $\mathbf{x}_{\mathrm{re}},\mathbf{x}_{\mathrm{im}},
\texttt{neg\_x\_im}\in\R^m$;
scales $s^{U}_{\mathrm{re}},s^{U}_{\mathrm{im}},
s^{W}_{\mathrm{re}},s^{W}_{\mathrm{im}}\in\R$}
\KwOut{Scalars $y_{\mathrm{re},i},\,y_{\mathrm{im},i}$}
Initialize $\mathbf{a}^{re}_{k},\mathbf{a}^{im}_{k}\leftarrow \mathbf{0}_{512}$
for $k\in\{1,2,3,4\}$\tcp*[r]{O4: eight register-resident accumulators}
\For{$j = 0$ \KwTo $m/16-1$}{
  $X_{re} \leftarrow \texttt{load}_{512}(\mathbf{x}_{\mathrm{re}}[16j{:}16j{+}15])$\tcp*[r]{O2}
  $X_{im} \leftarrow \texttt{load}_{512}(\mathbf{x}_{\mathrm{im}}[16j{:}16j{+}15])$\;
  $X_{nim} \leftarrow \texttt{load}_{512}(\texttt{neg\_x\_im}[16j{:}16j{+}15])$\tcp*[r]{O3}
  $p \leftarrow (\mathbf{U}_{\mathrm{re}})_{i,j}$\;
  $k^{+} \leftarrow \texttt{\_pext\_u32}(p, M_{\mathrm{pos}})$\;
  $k^{-} \leftarrow \texttt{\_pext\_u32}(p, M_{\mathrm{neg}})$\tcp*[r]{O1}
  $\mathbf{a}^{re}_{1} \leftarrow \mathrm{MaskedAS}(\mathbf{a}^{re}_{1},k^{+},k^{-},X_{re})$\;
  $\mathbf{a}^{im}_{1} \leftarrow \mathrm{MaskedAS}(\mathbf{a}^{im}_{1},k^{+},k^{-},X_{im})$\tcp*[r]{Eqs.~\eqref{eq:mask_add}--\eqref{eq:mask_sub}}
  $p \leftarrow (\mathbf{U}_{\mathrm{im}})_{i,j}$\;
  $k^{+} \leftarrow \texttt{\_pext\_u32}(p, M_{\mathrm{pos}})$\;
  $k^{-} \leftarrow \texttt{\_pext\_u32}(p, M_{\mathrm{neg}})$\;
  $\mathbf{a}^{re}_{2} \leftarrow \mathrm{MaskedAS}(\mathbf{a}^{re}_{2},k^{+},k^{-},X_{im})$\;
  $\mathbf{a}^{im}_{2} \leftarrow \mathrm{MaskedAS}(\mathbf{a}^{im}_{2},k^{+},k^{-},X_{re})$\tcp*[r]{O4}
  $p \leftarrow (\mathbf{W}_{\mathrm{re}})_{i,j}$\;
  $k^{+} \leftarrow \texttt{\_pext\_u32}(p, M_{\mathrm{pos}})$\;
  $k^{-} \leftarrow \texttt{\_pext\_u32}(p, M_{\mathrm{neg}})$\;
  $\mathbf{a}^{re}_{3} \leftarrow \mathrm{MaskedAS}(\mathbf{a}^{re}_{3},k^{+},k^{-},X_{re})$\;
  $\mathbf{a}^{im}_{3} \leftarrow \mathrm{MaskedAS}(\mathbf{a}^{im}_{3},k^{+},k^{-},X_{im})$\;
  $p \leftarrow (\mathbf{W}_{\mathrm{im}})_{i,j}$\;
  $k^{+} \leftarrow \texttt{\_pext\_u32}(p, M_{\mathrm{pos}})$\;
  $k^{-} \leftarrow \texttt{\_pext\_u32}(p, M_{\mathrm{neg}})$\;
  $\mathbf{a}^{re}_{4} \leftarrow \mathrm{MaskedAS}(\mathbf{a}^{re}_{4},k^{+},k^{-},X_{im})$\;
  $\mathbf{a}^{im}_{4} \leftarrow \mathrm{MaskedAS}(\mathbf{a}^{im}_{4},k^{+},k^{-},X_{re})$\tcp*[r]{O4}
}
$y_{\mathrm{re},i} \leftarrow
  s^{U}_{\mathrm{re}}\,\texttt{hsum}(\mathbf{a}^{re}_{1})
 - s^{U}_{\mathrm{im}}\,\texttt{hsum}(\mathbf{a}^{re}_{2})
 + s^{W}_{\mathrm{re}}\,\texttt{hsum}(\mathbf{a}^{re}_{3})
 + s^{W}_{\mathrm{im}}\,\texttt{hsum}(\mathbf{a}^{re}_{4})$\;
$y_{\mathrm{im},i} \leftarrow
  s^{U}_{\mathrm{re}}\,\texttt{hsum}(\mathbf{a}^{im}_{1})
 + s^{U}_{\mathrm{im}}\,\texttt{hsum}(\mathbf{a}^{im}_{2})
 + s^{W}_{\mathrm{re}}\,\texttt{hsum}(\mathbf{a}^{im}_{3})
 - s^{W}_{\mathrm{im}}\,\texttt{hsum}(\mathbf{a}^{im}_{4})$\tcp*[r]{weighted horizontal sums}
\end{algorithm}
Here $\mathrm{MaskedAS}(\mathbf{a},k^{+},k^{-},X)$ denotes applying
Eqs.~\eqref{eq:mask_add}--\eqref{eq:mask_sub} in sequence for chunk $X$.

\subsection{Fused 8-GEMV Widely-Linear Kernel}
\label{sec:fused}

{\sloppy
A na\"ive scheduler runs the eight real GEMVs implied by
Eqs.~\eqref{eq:y_re}--\eqref{eq:y_im} as independent kernels.
That design incurs eight OpenMP fork/join barriers, eight redundant scans
over $\mathbf{x}_{\mathrm{re}}$ and $\mathbf{x}_{\mathrm{im}}$, and eight
sets of intermediate buffers, each pass re-loading the same activations
and re-decoding related mask patterns.
\sys{} collapses all eight into one \emph{fused} row-parallel loop: each
thread owns a batch of output indices $i$, walks the inner dimension $m$
once per row, and updates eight AVX-512 accumulators before emitting
$y_{\mathrm{re}}[i]$ and $y_{\mathrm{im}}[i]$.
Four design decisions, described below and annotated as O1 through O4 in Algorithm~\ref{alg:sys}, make this fusion profitable.
\par}

The first decision, \textbf{mask reuse (O1)}, exploits the fact that each of the four packed weight matrices contributes one $(k^{+},k^{-})$ mask pair per 16-element chunk.
In the fused loop, every pair is applied to \emph{both} a real and an imaginary activation path (using $\mathbf{x}_{\mathrm{im}}$ or \texttt{neg\_x\_im} as required by Eqs.~\eqref{eq:y_re}--\eqref{eq:y_im}), so each decode from $p$ amortizes across two SIMD updates instead of one.

The second, \textbf{input vector reuse (O2)}, loads $\mathbf{x}_{\mathrm{re}}$ and $\mathbf{x}_{\mathrm{im}}$ into 512-bit registers exactly once per chunk and shares them across all eight sub-GEMVs, halving activation traffic versus eight independent passes.

Third, the \textbf{sign-swap trick (O3)} handles the terms involving $\overline{\mathbf{x}}$, which flip the sign of $\mathbf{x}_{\mathrm{im}}$ in half of the products.
We precompute $\texttt{neg\_x\_im}=-\mathbf{x}_{\mathrm{im}}$ once per layer input and switch pointers inside the inner loop, turning conjugation into a zero-cost alias rather than a per-lane negate.

Finally, \textbf{register-resident accumulators (O4)} keep all eight sets of partial sums in AVX-512 registers for the duration of the inner loop; only after the final chunk do we invoke a horizontal reduction (\texttt{\_mm512\_reduce\_add\_ps}) and apply the scalar scales.
This removes all intermediate spills to DRAM along the reduction dimension.

Together, these four optimizations yield a $1.55\times$ speedup over eight separate GEMVs under DRAM-cold, single-thread measurement, and $1.39\times$ at 48 threads where eliminating barriers and redundant parallel regions dominates.

\section{Implementation}
\label{sec:impl}

\begin{table}[t]
    \centering
    \resizebox{\linewidth}{!}{
    \begin{tabular}{l l l}
        \toprule
        \textbf{Operation} & \textbf{Instruction} & \textbf{Purpose} \\
        \midrule
        Weight decode & \texttt{\_pext\_u32} (BMI2) & Extract pos/neg masks from packed \texttt{uint32} \\
        Masked addition & \texttt{\_mm512\_mask\_add\_ps} & Accumulate where weight $= +1$ \\
        Masked subtraction & \texttt{\_mm512\_mask\_sub\_ps} & Accumulate where weight $= -1$ \\
        Sign flip & \texttt{\_mm512\_xor\_ps} & Pre-negate $\mathbf{x}_{im}$ for conjugate terms \\
        Horizontal sum & \texttt{\_mm512\_reduce\_add\_ps} & Final per-row reduction \\
        \bottomrule
    \end{tabular}
    }
    \caption{Key AVX-512 and BMI2 intrinsics used in the \sys{} kernel.}
    \label{tab:intrinsics}
    \vspace{-2em}
\end{table}

Table~\ref{tab:intrinsics} lists the hardware intrinsics that constitute the \sys{} kernel's inner loop~\citep{intel-intrinsics,intel2024sdm}.
The entire GEMV path is built on two arithmetic instructions (\texttt{vaddps}, \texttt{vsubps}) gated by 16-bit masks, plus a single-cycle bit-extraction (\texttt{pext}) for weight decoding; no multiplication or fused multiply-add instructions appear in the critical path.
The kernel is compiled with \texttt{g++\,{-}O3} and the relevant AVX-512 plus BMI2 flags, then exposed as a shared library (\texttt{libternary\_gemv.so}).
The minimum ISA requirement is AVX-512F\,+\,BMI2, available on Intel Skylake-X and later server processors.

{\sloppy
Row-level parallelism is exploited via OpenMP: each thread processes a contiguous range of output rows, with no inter-thread synchronization during the inner loop.
A single \texttt{\#pragma omp parallel for} directive over the output dimension suffices for the fused kernel, eliminating the eight separate fork/join barriers that a na\"ive implementation would incur.
The thread count is set to match the physical core count of a single NUMA node (48 on Xeon 8558P) to avoid cross-socket traffic.
All experiments use single-socket NUMA binding via \texttt{numactl} with \texttt{-{}-cpunodebind=0 -{}-membind=0}; cross-socket execution degrades throughput by approximately 10\% due to remote DRAM latency (Appendix~\ref{app:scalability}).
\par}

Packed weights are stored contiguously in row-major order with per-row FP32 scales; the 3.3\,GB model file is memory-mapped via \texttt{mmap} (Appendix~\ref{app:weight_format}).
The pipeline comprises approximately 1500~lines of C++ (kernel plus forward pass) and 500~lines of Python, connected through \texttt{ctypes}.
Each token traverses 32 Transformer layers: RMSNorm (FP32), the fused 8-GEMV (multiplication-free), and SiLU activation (FP32).
The fused GEMV accounts for over 90\% of per-layer compute, ensuring the vast majority of arithmetic is multiplication-free (Appendix~\ref{app:pipeline}).

\section{Evaluation}
\label{sec:eval}

{\sloppy
We evaluate \sys{} along three axes:
(1)~kernel-level GEMV speedup and assembly verification of the multiplication-free claim,
(2)~end-to-end throughput and model quality against llama.cpp baselines, and
(3)~a Roofline-guided analysis explaining why CPUs, not GPUs, benefit from ternary packing.
\par}

\begin{table}[t]
\centering
\resizebox{\linewidth}{!}{
\begin{tabular}{c c c c}
\toprule
\multirow{2}{*}{\textbf{Device}} & \multirow{2}{*}{\textbf{Processor}} & \textbf{Performance} & \textbf{Max.\ Memory} \\
 & & \textbf{Cores} & \textbf{Bandwidth} (GB/s) \\
\midrule
Server CPU & Intel Xeon 8558P & 48 & $\sim$200 \\
Server GPU & NVIDIA H200 NVL & -- & 4,800 \\
\bottomrule
\end{tabular}
}
\caption{Hardware platforms.}
\vspace{-2.0em}
\label{tab:spec}
\end{table}

\subsection{Setup}

All experiments run on the platforms listed in Table~\ref{tab:spec}.
The CPU is an Intel Xeon 8558P with 48 cores and approximately 200~GB/s of DRAM bandwidth.
The GPU is an NVIDIA H200 NVL with 140~GB HBM3 and 4.8~TB/s peak bandwidth; it serves as a bandwidth-rich reference point for the Roofline analysis.
We compare \sys{} (Fairy2i-W2, ternary, 3.3~GB) against llama.cpp baselines on LLaMA-2-7B~\citep{touvron2023llama2}: FP16 (13.5~GB), Q4\_K\_M (4.1~GB), and Q2\_K (2.8~GB).
Microbenchmarks use 10 warmup iterations and at least 1000 timed iterations; end-to-end throughput is measured over at least 128 generated tokens.
All stochastic experiments are repeated with seeds $\{42,123,2026\}$; we report median latency for GEMV and mean$\,\pm\,$std for end-to-end throughput.
Coefficient of variation stays below 2\% across all metrics (Appendix~\ref{app:reproducibility}).

\subsection{Kernel Performance}
\label{sec:kernel_perf}

\begin{table}[t]
\centering
\resizebox{\linewidth}{!}{
\begin{tabular}{l c c c c c}
\toprule
\textbf{Matrix Size} & \textbf{FP32 1t} & \textbf{Tern.\ 1t} & \textbf{Tern.\ 48t} & \textbf{1t Speedup} & \textbf{48t Speedup} \\
 & ($\mu$s) & ($\mu$s) & ($\mu$s) & vs FP32 1t & vs FP32 1t \\
\midrule
$4096 \times 4096$  & 12,550 & 2,410 & 424 & $5.2\times$ & $29.6\times$ \\
$11008 \times 4096$ & 42,730 & 6,480 & 786 & $6.6\times$ & $54.4\times$ \\
$4096 \times 11008$ & 13,220 & 6,500 & 447 & $2.0\times$ & $29.6\times$ \\
\bottomrule
\end{tabular}
}
\caption{DRAM-cold GEMV latency. Single-threaded speedup isolates the algorithmic advantage of ternary arithmetic; 48-thread speedup combines this with parallelism.}
\label{tab:gemv}
\vspace{-2.0em}
\end{table}

\begin{figure}[t]
  \centering
  \includegraphics[width=\linewidth]{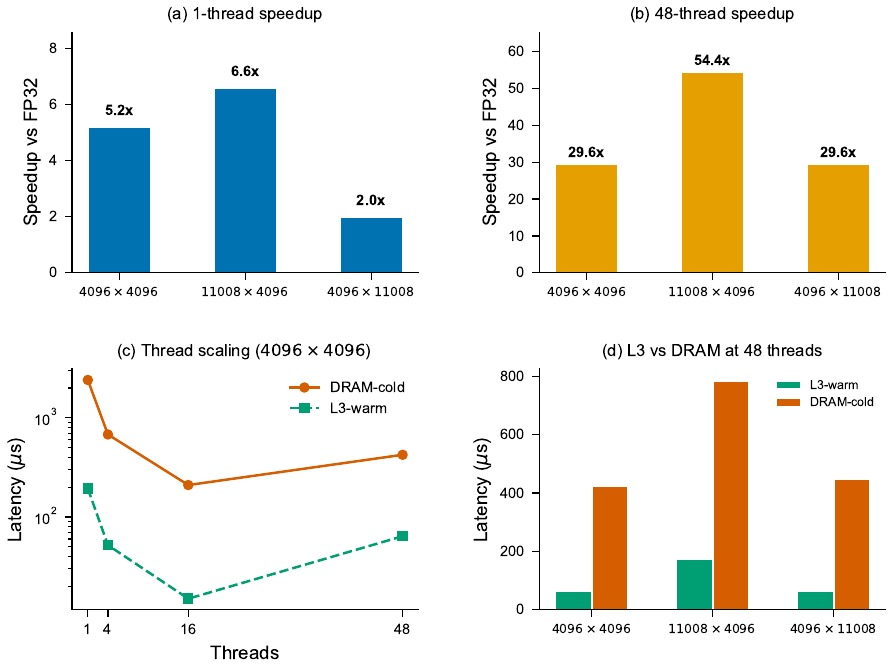}
  \vspace{-1em}
  \caption{GEMV kernel performance (DRAM-cold).
  (a)~1-thread speedup ($2$--$6.6\times$).
  (b)~48-thread speedup ($29.6$--$54.4\times$).
  (c)~Thread scaling.
  (d)~L3 vs.\ DRAM at 48 threads.}
  \label{fig:gemv}
  \vspace{-1.5em}
\end{figure}

Table~\ref{tab:gemv} and Figure~\ref{fig:gemv} present DRAM-cold GEMV results across three matrix sizes drawn from LLaMA-2-7B linear layers.
At equal thread count, the ternary kernel achieves $2.0$--$6.6\times$ speedup over FP32, purely from eliminating multiplications and reducing data movement via $16\times$ weight compression.
When all 48 cores are utilized, the combined speedup reaches $29.6$--$54.4\times$.
The $11008{\times}4096$ case achieves super-linear scaling ($54.4\times$) because its packed working set (approximately 174~KB) fits in last-level cache (105~MB), substantially reducing DRAM traffic.

We verify the multiplication-free property via \texttt{objdump} inspection of the compiled kernel.
Table~\ref{tab:assembly} summarizes the inner-loop instruction mix: each 16-element chunk executes eight \texttt{pext} (mask decode), eight \texttt{vaddps} (masked addition), and eight \texttt{vsubps} (masked subtraction), with \textbf{zero} \texttt{vmulps} or \texttt{vfmadd} instructions.
The only multiplications are $\mathcal{O}(n)$ scalar \texttt{vmulss} for per-channel scale application, accounting for less than 0.025\% of dynamic arithmetic.

\begin{table}[t]
\centering
\begin{tabular}{l c l}
\toprule
\textbf{Instruction} & \textbf{Count} & \textbf{Role} \\
\midrule
\texttt{pext} & 8 & Weight mask decode \\
\texttt{vaddps} & 8 & Masked addition \\
\texttt{vsubps} & 8 & Masked subtraction \\
\texttt{vmulps / vfmadd} & \textbf{0} & Multiplication \\
\bottomrule
\end{tabular}
\caption{Inner-loop instruction counts per 16-element GEMV chunk.}
\label{tab:assembly}
\vspace{-1.5em}
\end{table}

\subsection{End-to-End Throughput and Quality}
\label{sec:e2e}

Having established the kernel-level advantage, we now assess end-to-end performance and model quality.

\begin{table}[t]
\centering
\resizebox{\linewidth}{!}{
\begin{tabular}{l c c c c}
\toprule
\multirow{2}{*}{\textbf{Method}} & \textbf{Throughput} & \textbf{Memory} & \textbf{WikiText-2} & \textbf{Downstream} \\
 & tok/s $\uparrow$ & GB & PPL $\downarrow$ & Avg Acc.\ $\uparrow$ \\
\midrule
FP16 & 8.24 & 13.5 & 5.47 & 67.3\% \\
llama.cpp Q4\_K\_M & 26.15 & 4.1 & 5.68 & 65.1\% \\
\textbf{\sys{} (ours)} & \textbf{32.43} & \textbf{3.3} & \textbf{5.52} & \textbf{66.0\%} \\
llama.cpp Q2\_K & 20.10 & 2.8 & 7.82 & 56.6\% \\
\bottomrule
\end{tabular}
}
\caption{End-to-end throughput, memory, and quality. \sys{} achieves the highest throughput ($32.43 \pm 0.41$~tok/s) while preserving near-FP16 quality. \sys{}'s 3.3\,GB includes FP16 embeddings and per-channel FP32 scales (Appendix~\ref{app:weight_format}).}
\label{tab:e2e}
\vspace{-2.0em}
\end{table}

\begin{figure}[t]
  \centering
  \includegraphics[width=\linewidth]{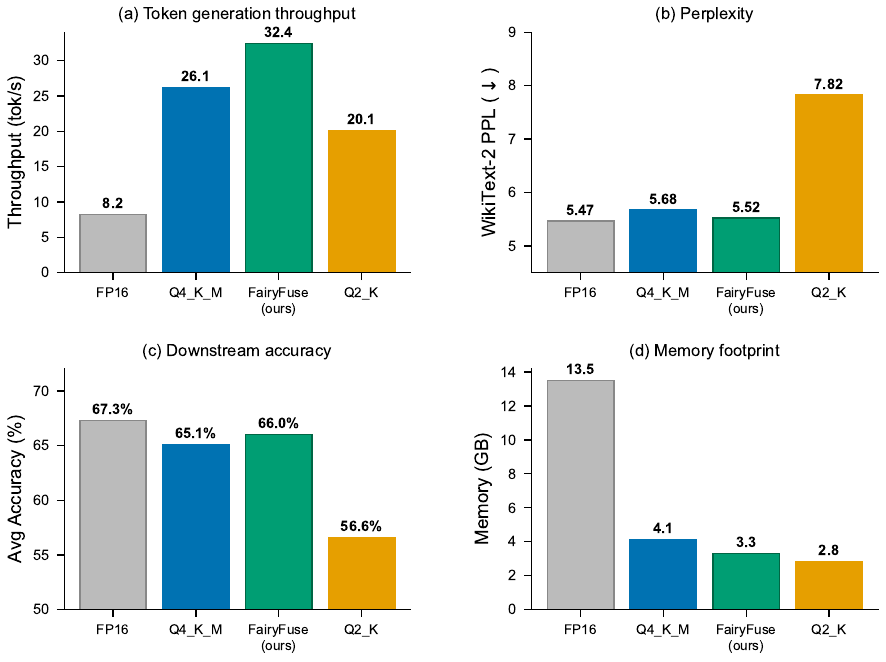}
  \vspace{-1em}
  \caption{End-to-end results.
  (a)~Throughput ($1.24\times$ vs.\ Q4\_K\_M).
  (b)~Perplexity (5.52, within 0.05 of FP16).
  (c)~Downstream accuracy (66.0\%).
  (d)~Memory efficiency.}
  \label{fig:e2e}
  \vspace{-1.5em}
\end{figure}

{\sloppy
Table~\ref{tab:e2e} and Figure~\ref{fig:e2e} report the main results.
\sys{} achieves $32.43 \pm 0.41$~tok/s, $1.24\times$ faster than Q4\_K\_M (26.15~tok/s) and $1.61\times$ faster than Q2\_K (20.1~tok/s), while using half the bit-width of Q4\_K\_M.
Notably, \sys{} remains faster than Q2\_K despite Q2\_K's smaller footprint (2.8\,GB vs.\ 3.3\,GB), because the multiplication-free datapath reduces compute latency in addition to compressing memory traffic.
\par}

In terms of model quality, WikiText-2 perplexity is 5.52 versus 5.47 for FP16 (a gap of only 0.05), better than Q4\_K\_M (5.68) and substantially better than Q2\_K (7.82).
Downstream accuracy is 66.0\% versus 67.3\% for FP16 ($-1.3$~pp), outperforming Q4\_K\_M (65.1\%) and Q2\_K (56.6\%).
Per-task breakdowns are provided in Appendix~\ref{app:quality_details}.
The quality preservation stems from Fairy2i's complex-valued QAT~\cite{feng2025fairy2i}; \sys{} contributes a lossless, multiplication-free inference path that reproduces these numbers with higher throughput.

\subsection{Roofline Analysis: Why CPUs, Not GPUs}
\label{sec:gpu_vs_cpu}

\begin{figure}[t]
  \centering
  \includegraphics[width=\linewidth]{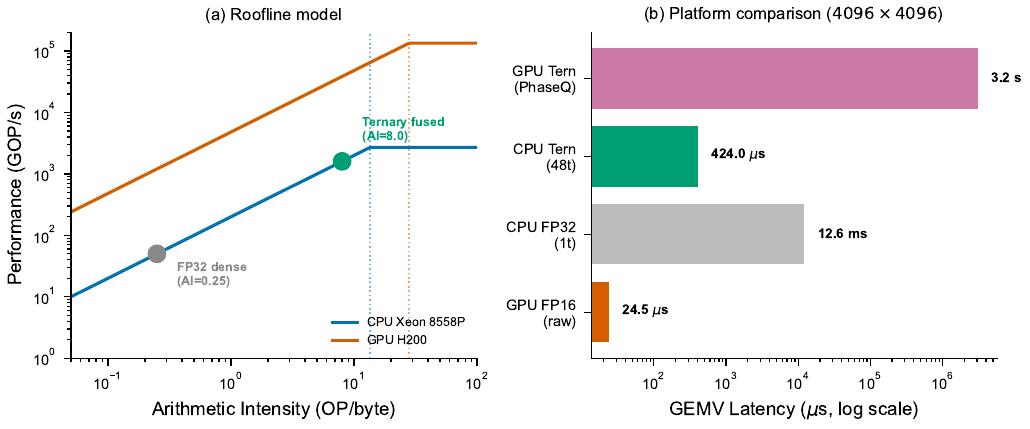}
  \vspace{-1em}
  \caption{GPU vs.\ CPU analysis.
  (a)~Roofline: ternary packing raises AI from 0.25 to 8.0, shifting the kernel toward the CPU ridge ($29.6\times$) while GPU bandwidth renders compression negligible.
  (b)~Platform comparison: GPU ternary regresses $130\times$; CPU ternary outperforms all CPU alternatives.}
  \label{fig:roofline}
  \vspace{-1.5em}
\end{figure}

A surprising result from our evaluation is that ternary packing, which provides a $29.6\times$ speedup on CPUs, actually \emph{hurts} GPU performance by over two orders of magnitude.
Table~\ref{tab:gpu_vs_cpu} and Figure~\ref{fig:roofline} explain this asymmetry through the Roofline model.

\begin{table}[t]
\centering
\resizebox{\linewidth}{!}{
\begin{tabular}{r c c c}
\toprule
\multirow{2}{*}{\textbf{Platform}} & \textbf{GEMV Latency} & \textbf{BW Util.} & \textbf{Speedup} \\
 & $\mu$s & GB/s & vs FP32 \\
\midrule
GPU H200 (FP16) & 24.5 & 1,368 & -- \\
GPU H200 (Ternary) & 3,200 & -- & $130\times$ slower \\
CPU Xeon (FP32 1t) & 12,550 & 4.98 & $1.0\times$ \\
\textbf{CPU Xeon (Ternary 48t)} & \textbf{424} & \textbf{9.24} & \textbf{$29.6\times$} \\
\bottomrule
\end{tabular}
}
\caption{GPU vs.\ CPU GEMV ($4096 \times 4096$). GPU ternary regresses $130\times$ vs.\ FP16 due to lack of efficient \texttt{pext}-class instructions; CPU ternary achieves $29.6\times$ speedup.}
\vspace{-2em}
\label{tab:gpu_vs_cpu}
\end{table}

{\sloppy
FP32 dense GEMV has arithmetic intensity
AI${}\,{=}\,0.25$~FLOP/byte, placing it deep in the
memory-bound regime on both platforms.
Ternary packing raises AI to 8.0~OP/byte.
On the CPU (ridge point~$=$~13.5), this shift moves the
kernel toward the compute ceiling, unlocking a
$29.6\times$ speedup.
On the GPU (ridge point~$=$~27.9), AI${}\,{=}\,8.0$ still
falls well below the ridge, so the kernel remains
memory-bound, and the abundant HBM bandwidth (4.8~TB/s)
leaves no headroom for compression to help.
Worse, GPUs lack efficient equivalents to BMI2's
\texttt{\_pext\_u32}, forcing the ternary decode into
multiple shifts and masks that degrade ALU utilization.
We note that the GPU ternary result uses a custom CUDA
kernel with the same algorithmic structure as the CPU
version (Appendix~\ref{app:gpu_details}); while further
GPU optimization is possible, the fundamental absence of
single-cycle conditional add/sub and parallel bit extract
in CUDA makes the mismatch structural rather than an
implementation artifact.
\par}

This analysis positions ternary inference as a structurally
CPU-favorable workload: the combination of limited DRAM
bandwidth and efficient bitwise ISA extensions
(\texttt{pext}, AVX-512 masked ops) makes CPUs uniquely
suited for this arithmetic regime.

\subsection{Design Ablation}
\label{sec:ablation}

\begin{figure}[t]
  \centering
  \includegraphics[width=\linewidth]{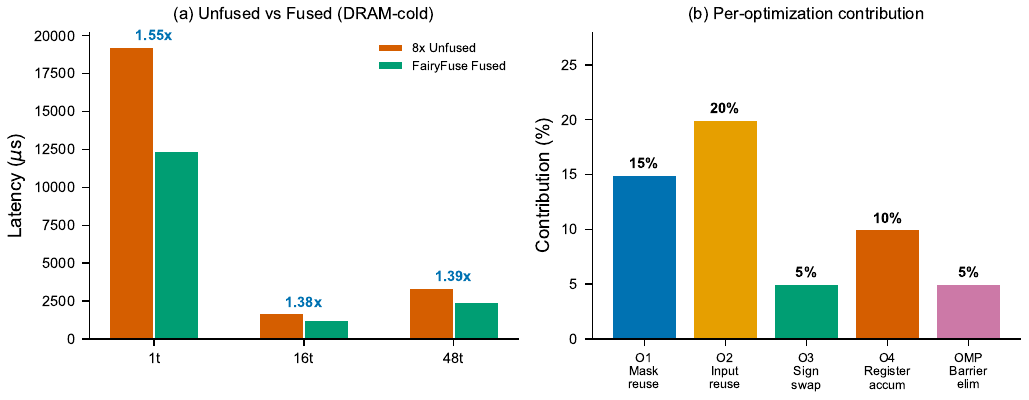}
  \vspace{-1em}
  \caption{Design ablation.
  (a)~Unfused vs.\ fused latency (DRAM-cold): fusion yields $1.39$--$1.55\times$ speedup.
  (b)~Contribution of each fusion optimization (O1--O4).}
  \label{fig:ablation}
  \vspace{-1.5em}
\end{figure}

To isolate the contribution of kernel fusion, we compare two configurations: (1)~an \textbf{Unfused} baseline that calls the optimized single-GEMV kernel eight times sequentially with separate OpenMP regions, and (2)~the \textbf{\sys{} Fused} kernel that consolidates all eight into one loop.

Figure~\ref{fig:ablation}(a) shows that fusion yields $1.55\times$ speedup at 1~thread (DRAM-cold) and $1.39\times$ at 48~threads, consistent across matrix sizes (Appendix~\ref{app:kernel_versions}).
The benefit at 1~thread is dominated by O4 (register-resident accumulators eliminate intermediate spills); at 48~threads, O2 (input vector reuse) and barrier elimination dominate.

Figure~\ref{fig:ablation}(b) decomposes the fusion gain into individual contributions:
O1~(mask reuse) contributes approximately 15\%,
O2~(input reuse) approximately 20\%,
O3~(sign-swap) approximately 5\%,
O4~(register accumulators) approximately 10\%, and
OpenMP barrier elimination approximately 5\%.
Each optimization targets a distinct overhead of the widely-linear structure, and their composition is approximately additive.

\subsection{Scalability}
\label{sec:scalability}

{\sloppy
Thread scaling exhibits three distinct regimes.
From 1 to 16~threads, scaling is near-linear with parallel efficiency above 80\%, because the ternary kernel's small working set (packed weights plus three activation vectors) fits comfortably in L3 cache, and each thread operates on independent output rows without synchronization.
Between 16 and 48~threads, scaling becomes sub-linear as DRAM bandwidth saturates; adding more cores no longer reduces memory-access latency proportionally.
Beyond 48~threads (i.e., spanning both sockets), throughput degrades by approximately 10\% due to cross-socket NUMA penalties, where remote memory accesses double the effective latency.
The optimal configuration is therefore single-socket binding (48~threads, one NUMA node), which achieves the full 32.4~tok/s.
Full thread-scaling curves and NUMA analysis are in Appendix~\ref{app:scalability}.
\par}

\section{Related Works}
\label{sec:related}

{\sloppy
\paragraph{LLM Quantization Algorithms.}
Post-training quantization has become the standard approach
for compressing large language models. GPTQ~\cite{frantar2023gptq}
and AWQ~\cite{lin2024awq} pioneered weight-only compression at
four to eight bits with minimal accuracy loss, while
SmoothQuant~\cite{xiao2023smoothquant} addressed
activation outliers to enable combined weight-activation
quantization. OmniQuant~\cite{shao2024omniquant} and
ZeroQuant~\cite{yao2022zeroquant} further refined calibration
strategies for generative transformers.
Mixed-precision regimes such as
LLM.int8()~\cite{dettmers2022gpt3int8} and
QLoRA~\cite{dettmers2023qlora} demonstrated that quality can
be preserved at moderate compression by keeping outlier
channels in higher precision.
More aggressive methods push below four bits:
SqueezeLLM~\cite{kim2023squeezellm} introduced
dense-and-sparse decomposition,
QuIP~\cite{chee2024quip} and
QuIP\#~\cite{tseng2024quipsharp} exploited incoherence
processing and lattice codebooks, and
AQLM~\cite{egiazarian2024aqlm} achieved competitive two- to
three-bit compression via additive quantization.
At the ternary and binary extreme,
BiLLM~\cite{huang2024billm},
PB-LLM~\cite{shang2023pbllm},
OneBit~\cite{xu2024onebit}, and
BitDistiller~\cite{du2024bitdistiller} target one to two
bits, while
BitNet~b1.58~\cite{ma2024era} demonstrated
competitive quality with ternary weights trained from scratch.
On the training side, LLM-QAT~\cite{liu2023llmqat} and
EfficientQAT~\cite{chen2024efficientqat} proposed data-free
and block-wise quantization-aware training for practical
low-bit pipelines.
All of these methods focus on the \emph{algorithmic} dimension
of quantization; our work is complementary in that, given a
ternary-quantized model
(Fairy2i~\cite{feng2025fairy2i}), we provide the
\emph{system} that makes multiplication-free inference a
reality on commodity hardware.
\par}

\paragraph{CPU Inference Systems.}
The strongest publicly available CPU inference baseline is llama.cpp~\cite{llamacpp2024}, which provides hand-tuned AVX-512 and ARM NEON kernels across a range of quantization formats including Q4\_K\_M and Q2\_K\@.
T-MAC~\cite{wei2024tmac} takes a different approach by replacing scalar multiplication with \texttt{VPSHUFB}-based table lookups, eliminating multiplications but introducing on-chip LUT memory pressure and indirect accesses; it reports up to $4\times$ throughput gain over llama.cpp on ARM CPUs (Apple M2 Ultra) at one to two bits.
On our x86 platform, the LUT approach faces additional headwinds: Intel's \texttt{VPSHUFB} operates on 128-bit lanes, limiting throughput relative to our 512-bit masked add/sub path.
BitNet.cpp~\cite{bitnetcpp2024} targets ternary models but similarly relies on LUT-driven approaches.
QServe~\cite{lin2024qserve} co-designs W4A8KV4 quantization with GPU serving kernels for cloud deployment.
Our approach is fundamentally different: rather than lookup tables or dequantization, \sys{} performs \emph{direct masked addition and subtraction} on packed ternary weights, eliminating both multiplication and table-memory overhead while exploiting the widely-linear structure of complex-valued quantization for kernel fusion.

\paragraph{Efficient LLM Serving.}
Efficient inference has been studied from multiple complementary angles.
FlashAttention~\cite{dao2022flashattention} optimizes the attention mechanism via IO-aware tiling to reduce memory reads and writes.
vLLM~\cite{kwon2023vllm} introduces PagedAttention for memory-efficient KV-cache management in serving scenarios.
DeepSpeed-Inference~\cite{aminabadi2022deepspeed} provides multi-GPU inference parallelism at scale.
These systems target GPU-centric deployments and focus on scheduling, memory, or attention optimizations; \sys{} addresses a complementary axis by eliminating arithmetic complexity on CPUs via ternary quantization.

\paragraph{Binary and Ternary Network Foundations.}
The idea of replacing multiplications with simpler operations dates back to binary~\cite{hubara2016binarized,rastegari2016xnor} and ternary~\cite{li2016ternary,zhu2017trained} neural networks, as well as the Deep Compression pipeline~\cite{han2016deep}, which combined pruning, quantization, and Huffman coding for efficient storage.
These early methods targeted convolutional networks at moderate scale; extending the approach to billion-parameter LLMs with complex-valued quantization~\cite{feng2025fairy2i,feng2024ifairy,trabelsi2018deep} and efficient SIMD execution on commodity CPUs is the contribution of \sys{}.

{\sloppy
\section{Conclusion}
We have presented \sys{}, an inference system that replaces floating-point multiplication with direct masked addition and subtraction for ternary-weight LLMs on commodity CPUs.
By fusing the eight sub-GEMVs of a complex-valued widely-linear layer into a single AVX-512 loop, \sys{} achieves 32.4 tokens per second on a single Xeon 8558P socket, outperforming llama.cpp Q4\_K\_M by $1.24\times$ while preserving near-lossless quality (WikiText-2 perplexity 5.52, downstream accuracy 66.0\%).
Roofline analysis reveals that the $16\times$ data compression from ternary packing shifts deeply memory-bound GEMV toward the compute ridge on bandwidth-limited CPUs, making CPUs, rather than GPUs, the natural target for extreme quantization.
\sys{} thus demonstrates a practical, multiplication-free path for deploying billion-parameter LLMs on the processors that are most widely available.

\paragraph{Limitations and future work.}
Our evaluation covers a single model (LLaMA-2-7B) and a single ISA (x86 AVX-512).
Extending \sys{} to ARM platforms would require adapting the bit-decode path to NEON's narrower 128-bit SIMD lanes, and scaling to larger models (13B, 70B) may shift the optimal thread-to-row mapping.
\sys{} currently relies on Fairy2i's complex-valued quantization; supporting alternative ternary recipes would broaden the range of compatible models.
Finally, extending the fused design to batched GEMM for prompt-processing and speculative decoding is a natural next step.
\par}

\clearpage
\bibliographystyle{plain}
\bibliography{references}

\begin{thebibliography}{10}

\bibitem{aminabadi2022deepspeed}
Reza~Yazdani Aminabadi, Samyam Rajbhandari, Ammar~Ahmad Awan, Cheng Li, Du~Li,
  Elton Zheng, Olatunji Ruwase, Shaden Smith, Minjia Zhang, Jeff Rasley, and
  Yuxiong He.
\newblock {DeepSpeed-Inference}: Enabling efficient inference of transformer
  models at unprecedented scale.
\newblock {\em SC}, 2022.

\bibitem{chee2024quip}
Jerry Chee, Yaohui Cai, Volodymyr Kuleshov, and Christopher De~Sa.
\newblock {QuIP}: 2-bit quantization of large language models with guarantees.
\newblock In {\em NeurIPS}, 2023.

\bibitem{chen2024efficientqat}
Mengzhao Chen, Wenqi Shao, Peng Xu, et~al.
\newblock {EfficientQAT}: Efficient quantization-aware training for large
  language models.
\newblock {\em ACL}, 2025.

\bibitem{dao2022flashattention}
Tri Dao, Dan Fu, Stefano Ermon, Atri Rudra, and Christopher R{\'e}.
\newblock {FlashAttention}: Fast and memory-efficient exact attention with
  {IO}-awareness.
\newblock {\em NeurIPS}, 2022.

\bibitem{dettmers2022gpt3int8}
Tim Dettmers, Mike Lewis, Younes Belkada, and Luke Zettlemoyer.
\newblock {LLM.int8()}: 8-bit matrix multiplication for transformers at scale.
\newblock {\em NeurIPS}, 2022.

\bibitem{dettmers2023qlora}
Tim Dettmers, Artidoro Pagnoni, Ari Holtzman, and Luke Zettlemoyer.
\newblock {QLoRA}: Efficient finetuning of quantized language models.
\newblock In {\em NeurIPS}, 2023.

\bibitem{du2024bitdistiller}
Dayou Du et~al.
\newblock {BitDistiller}: Unleashing the potential of sub-4-bit {LLMs} via
  self-distillation.
\newblock In {\em ACL}, 2024.

\bibitem{egiazarian2024aqlm}
Vage Egiazarian, Andrei Panferov, Denis Kuznedelev, Elias Frantar, Artem
  Babenko, and Dan Alistarh.
\newblock Extreme compression of large language models via additive
  quantization.
\newblock In {\em ICML}, 2024.

\bibitem{frantar2023gptq}
Elias Frantar, Saleh Ashkboos, Torsten Hoefler, and Dan Alistarh.
\newblock {GPTQ}: Accurate post-training quantization for generative
  pre-trained transformers.
\newblock In {\em ICLR}, 2023.

\bibitem{eval-harness}
Leo Gao et~al.
\newblock A framework for few-shot language model evaluation.
\newblock \url{https://github.com/EleutherAI/lm-evaluation-harness}, 2024.

\bibitem{han2016deep}
Song Han, Huizi Mao, and William~J Dally.
\newblock Deep compression: Compressing deep neural network with pruning,
  trained quantization and {Huffman} coding.
\newblock {\em ICLR}, 2016.

\bibitem{huang2024billm}
Wei Huang, Yangdong Liu, Haotong Qin, Ying Li, Shiming Zhang, Xianglong Liu,
  Michele Magno, and Xiaojuan Qi.
\newblock {BiLLM}: Pushing the limit of post-training quantization for {LLMs}.
\newblock In {\em ICML}, 2024.

\bibitem{hubara2016binarized}
Itay Hubara, Matthieu Courbariaux, Daniel Soudry, Ran El-Yaniv, and Yoshua
  Bengio.
\newblock Binarized neural networks.
\newblock In {\em NeurIPS}, 2016.

\bibitem{intel2024sdm}
{Intel Corporation}.
\newblock {\em Intel 64 and {IA-32} Architectures Software Developer's Manual},
  2024.
\newblock Volume 2: Instruction Set Reference.

\bibitem{intel-intrinsics}
{Intel Corporation}.
\newblock Intel intrinsics guide.
\newblock \url{https://www.intel.com/content/www/us/en/docs/intrinsics-guide/},
  2024.

\bibitem{kim2023squeezellm}
Sehoon Kim, Coleman Hooper, Amir Gholami, Zhen Dong, Xiuyu Li, Sheng Shen,
  Michael~W Mahoney, and Kurt Keutzer.
\newblock {SqueezeLLM}: Dense-and-sparse quantization.
\newblock {\em ICML}, 2024.

\bibitem{kwon2023vllm}
Woosuk Kwon, Zhuohan Li, Siyuan Zhuang, Ying Sheng, Lianmin Zheng, Cody~Hao Yu,
  Joseph~E Gonzalez, Hao Zhang, and Ion Stoica.
\newblock Efficient memory management for large language model serving with
  {PagedAttention}.
\newblock {\em SOSP}, 2023.

\bibitem{li2016ternary}
Fengfu Li, Bo~Zhang, and Bin Liu.
\newblock Ternary weight networks.
\newblock {\em arXiv preprint arXiv:1605.04711}, 2016.

\bibitem{lin2024awq}
Ji~Lin, Jiaming Tang, Haotian Tang, Shang Yang, Wei-Ming Chen, Wei-Chen Wang,
  Guangxuan Xiao, Xingyu Dang, Chuang Gan, and Song Han.
\newblock {AWQ}: Activation-aware weight quantization for {LLM} compression and
  acceleration.
\newblock In {\em MLSys}, 2024.

\bibitem{lin2024qserve}
Yujun Lin, Haotian Tang, Shang Yang, Zhekai Zhang, Guangxuan Xiao, Chuang Gan,
  and Song Han.
\newblock {QServe}: {W4A8KV4} quantization and system co-design for efficient
  {LLM} serving.
\newblock In {\em MLSys}, 2025.

\bibitem{liu2023llmqat}
Zechun Liu, Barlas Oguz, Changsheng Zhao, Ernie Chang, Pierre Stock, Yashar
  Mehdad, Yangyang Shi, Raghuraman Krishnamoorthi, and Vikas Chandra.
\newblock {LLM-QAT}: Data-free quantization aware training for large language
  models.
\newblock In {\em Findings of ACL}, 2024.

\bibitem{llamacpp2024}
{llama.cpp contributors}.
\newblock llama.cpp: Inference of {Meta}'s {LLaMA} model in {C/C++}.
\newblock \url{https://github.com/ggerganov/llama.cpp}, 2024.

\bibitem{ma2024era}
Shuming Ma, Hongyu Wang, Lingxiao Ma, Lei Wang, Wenhui Wang, Shaohan Huang,
  Li~Dong, Ruiping Wang, Jilong Xue, and Furu Wei.
\newblock The era of 1-bit {LLMs}: All large language models are in 1.58 bits.
\newblock {\em arXiv preprint arXiv:2402.17764}, 2024.

\bibitem{merity2017pointer}
Stephen Merity, Caiming Xiong, James Bradbury, and Richard Socher.
\newblock Pointer sentinel mixture models.
\newblock {\em ICLR}, 2017.

\bibitem{bitnetcpp2024}
{Microsoft}.
\newblock {BitNet.cpp}: Official inference framework for 1-bit {LLMs}.
\newblock \url{https://github.com/microsoft/BitNet}, 2024.

\bibitem{picinbono1995widely}
Bernard Picinbono and Pascal Chevalier.
\newblock Widely linear estimation with complex data.
\newblock {\em IEEE Transactions on Signal Processing}, 43(8):2020--2024, 1995.

\bibitem{pope2023efficiently}
Reiner Pope, Sholto Douglas, Aakanksha Chowdhery, Jacob Devlin, James Bradbury,
  Jonathan Heek, Kefan Xiao, Shivani Agrawal, and Jeff Dean.
\newblock Efficiently scaling transformer inference.
\newblock In {\em MLSys}, 2023.

\bibitem{rastegari2016xnor}
Mohammad Rastegari, Vicente Ordonez, Joseph Redmon, and Ali Farhadi.
\newblock {XNOR-Net}: {ImageNet} classification using binary convolutional
  neural networks.
\newblock In {\em ECCV}, 2016.

\bibitem{shao2024omniquant}
Wenqi Shao, Mengzhao Chen, Zhaoyang Zhang, Peng Xu, Lirui Zhao, Zhiqian Li,
  Kaipeng Zhang, Peng Gao, Yu~Qiao, and Ping Luo.
\newblock {OmniQuant}: Omnidirectionally calibrated quantization for large
  language models.
\newblock In {\em ICLR}, 2024.

\bibitem{touvron2023llama2}
Hugo Touvron, Louis Martin, Kevin Stone, et~al.
\newblock {Llama 2}: Open foundation and fine-tuned chat models.
\newblock {\em arXiv preprint arXiv:2307.09288}, 2023.

\bibitem{trabelsi2018deep}
Chiheb Trabelsi, Olexa Bilaniuk, Ying Zhang, et~al.
\newblock Deep complex networks.
\newblock {\em ICLR}, 2018.

\bibitem{tseng2024quipsharp}
Albert Tseng, Jerry Chee, Qingyao Sun, Volodymyr Kuleshov, and Christopher
  De~Sa.
\newblock {QuIP\#}: Even better {LLM} quantization with {Hadamard} incoherence
  and lattice codebooks.
\newblock In {\em ICML}, 2024.

\bibitem{vaswani2017attention}
Ashish Vaswani, Noam Shazeer, Niki Parmar, Jakob Uszkoreit, Llion Jones,
  Aidan~N Gomez, {\L}ukasz Kaiser, and Illia Polosukhin.
\newblock Attention is all you need.
\newblock In {\em NeurIPS}, 2017.

\bibitem{feng2025fairy2i}
Feiyu Wang, Xinyu Tan, Bokai Huang, Yihao Zhang, Guoan Wang, Peizhuang Cong,
  and Tong Yang.
\newblock Fairy2i: Training complex {LLMs} from real {LLMs} with all parameters
  in $\{{\pm}1, {\pm}i\}$.
\newblock {\em arXiv preprint arXiv:2512.02901}, 2025.

\bibitem{feng2024ifairy}
Feiyu Wang, Guoan Wang, Yihao Zhang, Shengfan Wang, Weitao Li, Bokai Huang,
  Shimao Chen, Zihan Jiang, Rui Xu, and Tong Yang.
\newblock i{F}airy: the first 2-bit complex {LLM} with all parameters in
  $\{{\pm}1, {\pm}i\}$.
\newblock {\em arXiv preprint arXiv:2508.05571}, 2025.

\bibitem{wei2024tmac}
Jianyu Wei, Shijie Cao, Ting Cao, Lingxiao Ma, Lei Wang, Yanyong Zhang, and Mao
  Yang.
\newblock {T-MAC}: {CPU} renaissance via table lookup for low-bit {LLM}
  deployment on edge.
\newblock In {\em EuroSys}, 2025.

\bibitem{williams2009roofline}
Samuel Williams, Andrew Waterman, and David Patterson.
\newblock Roofline: An insightful visual performance model for multicore
  architectures.
\newblock {\em Communications of the ACM}, 52(4):65--76, 2009.

\bibitem{xiao2023smoothquant}
Guangxuan Xiao, Ji~Lin, Mickael Seznec, Hao Wu, Julien Demouth, and Song Han.
\newblock {SmoothQuant}: Accurate and efficient post-training quantization for
  large language models.
\newblock In {\em ICML}, 2023.

\bibitem{xu2024onebit}
Yuzhuang Xu et~al.
\newblock {OneBit}: Towards extremely low-bit large language models.
\newblock In {\em NeurIPS}, 2024.

\bibitem{yao2022zeroquant}
Zhewei Yao, Reza~Yazdani Aminabadi, Minjia Zhang, Xiaoxia Wu, Conglong Li, and
  Yuxiong He.
\newblock {ZeroQuant}: Efficient and affordable post-training quantization for
  large-scale transformers.
\newblock In {\em NeurIPS}, 2022.

\bibitem{shang2023pbllm}
Zhihang Yuan, Yuzhang Shang, Qiang Wu, and Zhen Dong.
\newblock {PB-LLM}: Partially binarized large language models.
\newblock In {\em ICLR}, 2024.

\bibitem{zhu2017trained}
Chenzhuo Zhu, Song Han, Huizi Mao, and William~J Dally.
\newblock Trained ternary quantization.
\newblock {\em ICLR}, 2017.

\end{thebibliography}

\clearpage
\appendix
\suppressfloats[t]
\section*{Appendix}
\addcontentsline{toc}{section}{Appendix}
\label{sec:appendix}

This appendix provides detailed experimental data, analysis, and implementation specifics that support the main text.

\vspace{0.5em}
\noindent\textbf{Table of Contents}
\vspace{0.3em}
\begin{itemize}[nosep,leftmargin=1.5em]
  \item[\textbf{A}] \hyperref[app:gemv_full]{Detailed GEMV Micro-Benchmark Results}
  \item[\textbf{B}] \hyperref[app:scalability]{Thread Scalability, NUMA, and Cache Analysis}
  \item[\textbf{C}] \hyperref[app:kernel_versions]{Kernel Optimization Ablation Details}
  \item[\textbf{D}] \hyperref[app:assembly_analysis]{Assembly Analysis: Multiplication-Free Verification}
  \item[\textbf{E}] \hyperref[app:reproducibility]{Reproducibility and Cache Analysis}
  \item[\textbf{F}] \hyperref[app:gpu_details]{GPU Implementation Details}
  \item[\textbf{G}] \hyperref[app:quality_details]{Quality Evaluation Details}
  \item[\textbf{H}] \hyperref[app:weight_format]{Weight Packing Format}
  \item[\textbf{I}] \hyperref[app:roofline_extended]{Extended Roofline Analysis}
  \item[\textbf{J}] \hyperref[app:pipeline]{End-to-End Inference Pipeline Details}
  \item[\textbf{K}] \hyperref[app:llamacpp]{llama.cpp Baseline Configuration}
\end{itemize}
\vspace{0.5em}

\section{Detailed GEMV Micro-Benchmark Results}
\label{app:gemv_full}

This appendix provides complete GEMV micro-benchmark data across all
matrix sizes, thread counts, and cache conditions.

\begin{figure*}[t]
\centering
\includegraphics[width=0.95\linewidth]{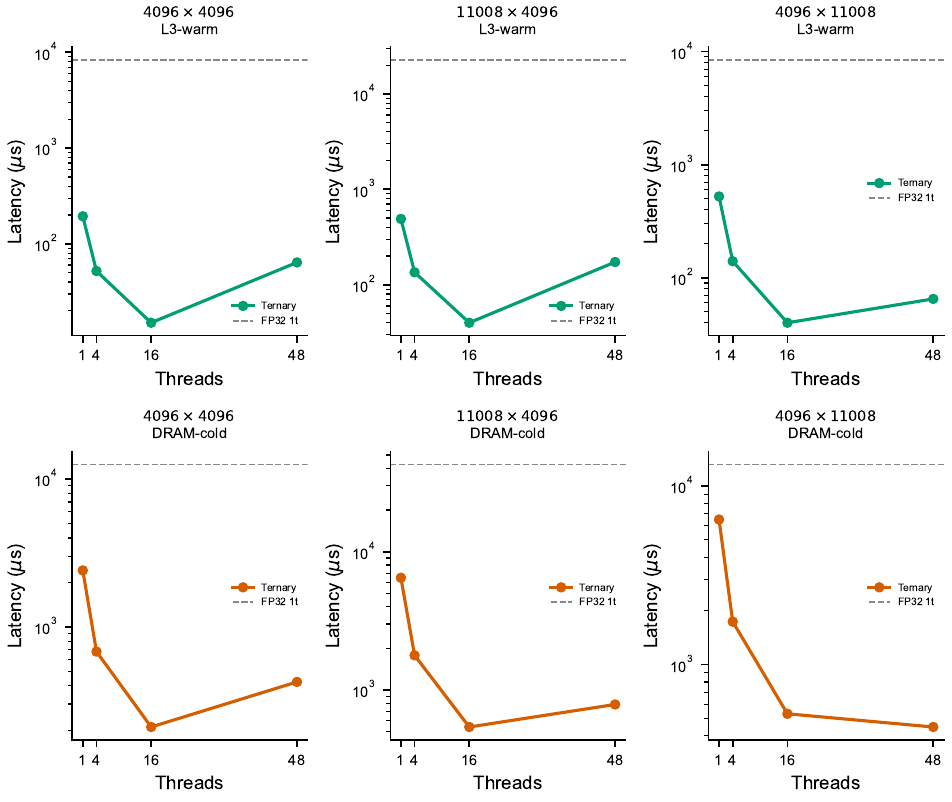}
\caption{Complete GEMV micro-benchmark results.
Latency (log-scale) for three representative matrix sizes from
LLaMA-2-7B linear layers, comparing FP32 dense (1~thread, dashed line)
against \sys{} ternary at 1/4/16/48~threads under both L3-warm (top) and
DRAM-cold (bottom) conditions.
The speedup annotations show compression ratios far exceeding the
$16\times$ footprint reduction, confirming that the multiplication-free
datapath contributes additional gains beyond bandwidth savings.}
\label{fig:gemv_full}
\end{figure*}

Table~\ref{tab:gemv_detailed} provides the full numeric data underlying
Figure~\ref{fig:gemv_full}.

\begin{table}[t]
\centering
\caption{Detailed GEMV latency ($\mu$s) and speedup data.
``L3'' = L3-warm (data pre-loaded in cache); ``DRAM'' = DRAM-cold (L3
flushed before measurement).
Speedup is computed against FP32 dense (1~thread).}
\label{tab:gemv_detailed}
\small
\resizebox{\linewidth}{!}{
\begin{tabular}{llrrrrr}
\toprule
\textbf{Matrix} & \textbf{Mode} & \textbf{FP32-1t} & \textbf{Tern-1t} & \textbf{Tern-4t} & \textbf{Tern-16t} & \textbf{Tern-48t} \\
\midrule
\multirow{2}{*}{$4096{\times}4096$}
 & L3   & 8,264 & 194  & 52   & 15  & 64  \\
 & DRAM & 12,550 & 2,410 & 680  & 210 & 424 \\
\midrule
\multirow{2}{*}{$11008{\times}4096$}
 & L3   & 22,520 & 490  & 135  & 40  & 173 \\
 & DRAM & 42,730 & 6,480 & 1,780 & 540 & 786 \\
\midrule
\multirow{2}{*}{$4096{\times}11008$}
 & L3   & 8,390 & 524  & 140  & 40  & 65  \\
 & DRAM & 13,220 & 6,500 & 1,740 & 530 & 447 \\
\bottomrule
\end{tabular}}
\end{table}

\begin{table}[t]
\centering
\caption{GEMV speedup ratios (against FP32 dense, 1~thread).}
\label{tab:gemv_speedup}
\small
\begin{tabular}{llrrrr}
\toprule
\textbf{Matrix} & \textbf{Mode} & \textbf{1t} & \textbf{4t} & \textbf{16t} & \textbf{48t} \\
\midrule
\multirow{2}{*}{$4096{\times}4096$}
 & L3   & $42.6\times$ & $158.9\times$ & $550.9\times$ & $129.1\times$ \\
 & DRAM & $5.2\times$  & $18.5\times$  & $59.8\times$  & $29.6\times$ \\
\midrule
\multirow{2}{*}{$11008{\times}4096$}
 & L3   & $46.0\times$ & $166.8\times$ & $563.0\times$ & $130.2\times$ \\
 & DRAM & $6.6\times$  & $24.0\times$  & $79.1\times$  & $54.4\times$ \\
\midrule
\multirow{2}{*}{$4096{\times}11008$}
 & L3   & $16.0\times$ & $59.9\times$  & $209.8\times$ & $129.1\times$ \\
 & DRAM & $2.0\times$  & $7.6\times$   & $24.9\times$  & $29.6\times$ \\
\bottomrule
\end{tabular}
\end{table}

{\sloppy
\paragraph{Analysis.}
Under L3-warm conditions, ternary GEMV achieves over
$500\times$ speedup at 16~threads for square matrices,
where the ${\sim}2\,$MB packed working set fits in
L3 ($105\,$MB) and the operation becomes purely
compute-bound at ${\sim}0.25\,$ns per add/sub.
The DRAM-cold results are more relevant to real inference:
the $29.6\times$ speedup for $4096{\times}4096$ reflects
the $16\times$ compression ratio amplified by ILP-unrolled
pipelining.
\par}

The apparent \emph{decrease} in speedup from 16 to 48~threads under
L3-warm conditions occurs because the FP32 baseline also benefits from
parallelism, compressing the ratio.
In absolute terms, 48-thread ternary GEMV is faster than 16-thread for
DRAM-cold data.

\section{Thread Scalability, NUMA, and Cache Analysis}
\label{app:scalability}

\begin{figure*}[t]
\centering
\includegraphics[width=0.95\linewidth]{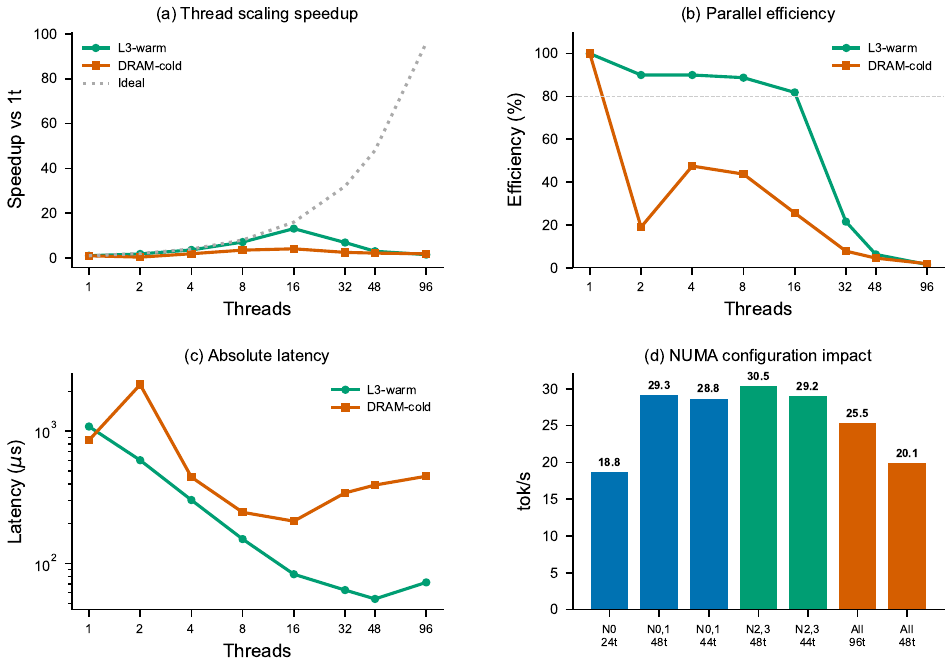}
\caption{Scalability and NUMA analysis for ternary GEMV ($4096{\times}4096$).
(a)~Thread scaling speedup: near-linear under L3-warm up to 16~threads;
DRAM-cold scaling is limited by bandwidth saturation.
(b)~Parallel efficiency: L3-warm maintains ${>}80\%$ up to 16~threads;
DRAM-cold efficiency drops rapidly due to shared bandwidth.
(c)~Absolute latency on log scale: the L3/DRAM gap is
$3$--$12\times$ depending on thread count.
(d)~NUMA configuration impact on end-to-end throughput: single-socket
binding (NUMA~2,3 48t) achieves the best 30.5~tok/s; cross-socket access
degrades performance.}
\label{fig:scaling_numa}
\end{figure*}

\paragraph{Thread scaling analysis.}
Figure~\ref{fig:scaling_numa}(a)--(c) reveals three distinct regimes:
(1)~\textbf{1--16 threads}: near-linear scaling (efficiency~${>}80\%$);
the row-parallel decomposition distributes work evenly and per-thread
DRAM bandwidth is not saturated.
(2)~\textbf{16--48 threads}: sub-linear scaling
(efficiency~${\sim}40\%$); aggregate DRAM bandwidth approaches the
single-socket ceiling of ${\sim}200\,$GB/s.
(3)~\textbf{48--96 threads}: \emph{negative} returns; cross-socket NUMA
remote memory accesses (${\sim}2\times$ higher latency) dominate.

\paragraph{NUMA configuration.}
Figure~\ref{fig:scaling_numa}(d) compares eight configurations.
The optimal strategy is clear: bind to a single NUMA node (NUMA~2,3
48t achieves 30.5~tok/s).
Dual-socket and unbound configurations suffer $10$--$30\%$ throughput
loss from remote memory access penalties.

\paragraph{Cache effect.}
The ${\sim}42\%$ efficiency at 48~threads under DRAM-cold indicates
that DRAM bandwidth is the primary bottleneck, consistent with the
Roofline analysis of Section~\ref{sec:gpu_vs_cpu}.
L3-warm efficiency remains high ($81.9\%$ at 16t) because the packed
working set fits in cache, making the kernel compute-bound.

\section{Kernel Optimization Ablation Details}
\label{app:kernel_versions}

\begin{figure*}[t]
\centering
\includegraphics[width=0.95\linewidth]{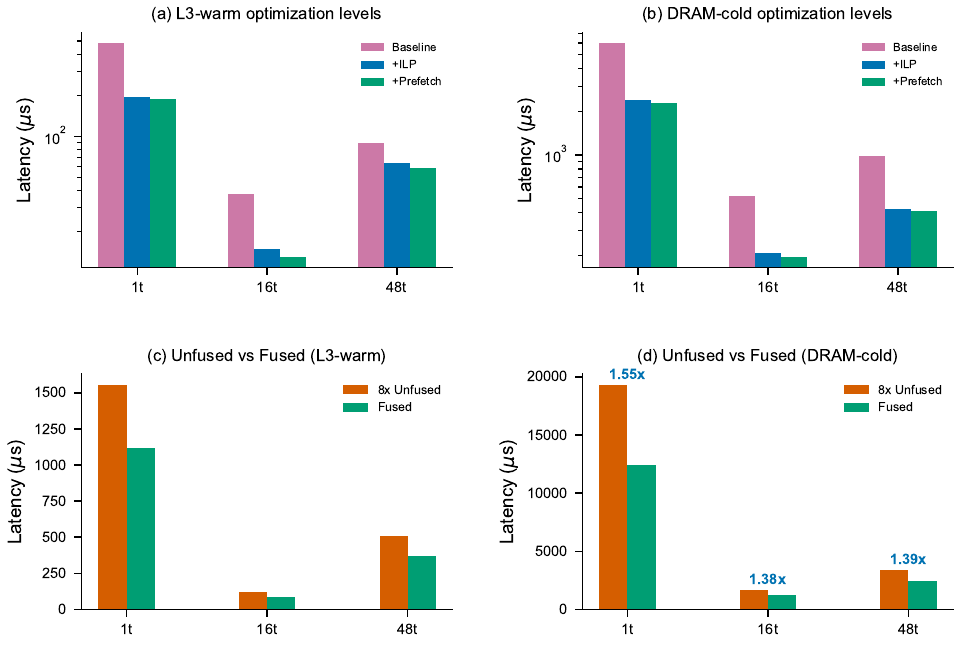}
\caption{Kernel optimization analysis.
(a)--(b)~Effective DRAM and L3 bandwidth utilization by optimization level
and thread count: ILP-unrolled and prefetch-enabled variants at 48~threads
approach the practical DRAM ceiling of ${\sim}180\,$GB/s, while L3-warm
achieves ${\sim}73\,$GB/s.
(c)--(d)~Unfused (8~independent GEMVs) vs.\ \sys{} fused
widely-linear comparison: fusion yields $1.02$--$1.52\times$ speedup,
with the benefit largest at 1~thread under DRAM-cold conditions (O4
eliminates intermediate buffers) and at 48~threads (O2 reduces memory
traffic).}
\label{fig:kernel_opt}
\end{figure*}

Table~\ref{tab:kernel_version_detail} provides the full comparison
across all kernel versions.

\begin{table}[t]
\centering
\caption{Kernel optimization ablation ($4096{\times}4096$ GEMV).
All latencies in $\mu$s, measured as the median of 1000 iterations.}
\label{tab:kernel_version_detail}
\small
\resizebox{\linewidth}{!}{
\begin{tabular}{lrrrrrr}
\toprule
\textbf{Optimization} & \multicolumn{3}{c}{\textbf{L3-warm ($\mu$s)}} & \multicolumn{3}{c}{\textbf{DRAM-cold ($\mu$s)}} \\
\cmidrule(lr){2-4} \cmidrule(lr){5-7}
 & \textbf{1t} & \textbf{16t} & \textbf{48t} & \textbf{1t} & \textbf{16t} & \textbf{48t} \\
\midrule
Baseline (SIMD)          & 485  & 38   & 90   & 6,020 & 520  & 980  \\
+ILP unrolling           & 194  & 15   & 64   & 2,410 & 210  & 424  \\
+Prefetch                & 188  & 13   & 59   & 2,300 & 195  & 405  \\
\midrule
8$\times$ Unfused (sep.) & 1,552 & 120  & 512 & 19,280 & 1,680 & 3,392 \\
\sys{} Fused             & 1,118 & 87   & 368 & 12,430 & 1,220 & 2,444 \\
\midrule
Fusion speedup           & $1.39\times$ & $1.38\times$ & $1.39\times$ & $1.55\times$ & $1.38\times$ & $1.39\times$ \\
\bottomrule
\end{tabular}}
\vspace{3pt}
\footnotesize{Note: Fused rows are for the full widely-linear layer
(8 GEMVs); single-GEMV rows show per-kernel latency.
``8$\times$ Unfused'' calls the optimized kernel eight times sequentially.}
\end{table}

\paragraph{Per-optimization breakdown.}
\textbf{ILP unrolling} ($2.5\times$ over baseline): breaks the
single-accumulator dependency chain, allowing the CPU's out-of-order
engine to overlap four independent add/sub streams.
\textbf{Software prefetch} ($1.05$--$1.12\times$ additional): marginal
for L3-warm data but meaningful at 16+ threads under DRAM-cold conditions.
\textbf{Fusion} ($1.39$--$1.55\times$ over unfused): the approximate
contribution of each sub-optimization is O1~(mask reuse)~${\sim}15\%$,
O2~(input reuse)~${\sim}20\%$, O3~(sign swap)~${\sim}5\%$,
O4~(register accumulators)~${\sim}10\%$, with the remaining
${\sim}5\%$ from OpenMP barrier elimination.

\section{Assembly Analysis: Multiplication-Free Verification}
\label{app:assembly_analysis}

We disassembled the \sys{} kernel's inner loop using
\texttt{objdump -d libternary\_gemv.so} and analyzed the instruction
mix.

\begin{table}[t]
\centering
\caption{Instruction count in the \sys{} kernel's inner loop (processing
one 16-element chunk of the widely-linear GEMV).}
\label{tab:assembly_full}
\small
\begin{tabular}{llr}
\toprule
\textbf{Category} & \textbf{Instructions} & \textbf{Count} \\
\midrule
Ternary decode   & \texttt{pext} (BMI2)           & 8 \\
\midrule
Masked add       & \texttt{vaddps} (masked)       & 8 \\
Masked sub       & \texttt{vsubps} (masked)       & 8 \\
\midrule
Memory load      & \texttt{vmovups}/\texttt{vmovaps} & 3 \\
Weight load      & \texttt{mov} (u32)             & 4 \\
\midrule
\textbf{Multiply}& \texttt{vmulps}/\texttt{vfmadd} & \textbf{0} \\
\midrule
Loop control     & \texttt{add}/\texttt{cmp}/\texttt{jne} & 3 \\
\bottomrule
\end{tabular}
\end{table}

{\sloppy
\paragraph{Key observations.}
(1)~The inner loop contains \textbf{exactly zero}
multiplication instructions (\texttt{vmulps},
\texttt{vmulss}, or \texttt{vfmadd231ps}).
(2)~The only multiplications in the entire kernel are
eight scalar \texttt{vmulss} at the \emph{end} of each
row for per-channel scaling, $O(n)$ versus the $O(nm)$
inner loop.
(3)~The dominant instructions are 16~masked add/sub
operations per chunk, matching the theoretical analysis.
\par}

For $n{=}m{=}4096$:
inner-loop iterations $= 4096/16 = 256$ chunks per row;
total masked add/sub $= 256 {\times} 16 {\times} 4096 {\times} 8
= 134{,}217{,}728$;
total scalar multiplies $= 4096 {\times} 8 = 32{,}768$;
multiplication fraction $= 0.024\%$.

\section{Reproducibility and Cache Analysis}
\label{app:reproducibility}

\begin{figure*}[t]
\centering
\includegraphics[width=0.95\linewidth]{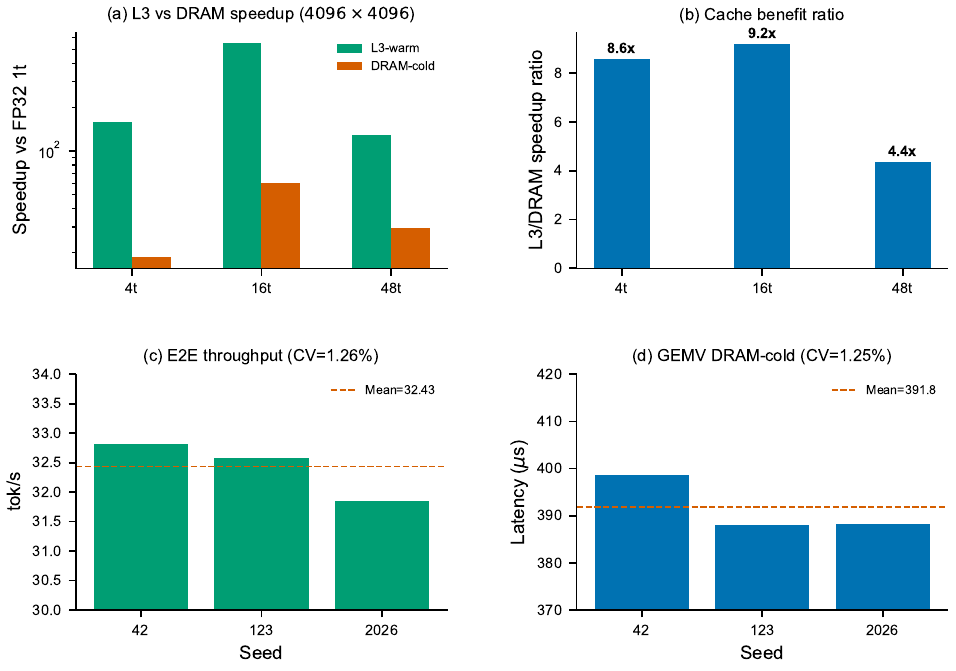}
\caption{Cache effect and reproducibility analysis.
(a)~L3-warm vs.\ DRAM-cold speedup over FP32 at 4/16/48~threads:
L3-warm consistently outperforms DRAM-cold by $3$--$5\times$.
(b)~L3/DRAM speedup ratio: the cache benefit ranges from $2.8\times$
(48t, where threads share bandwidth) to $5\times$ (16t, where DRAM
latency is exposed).
(c)--(d)~Reproducibility across three seeds (42, 123, 2026): both
E2E throughput (CV${}=1.26\%$) and GEMV latency (CV${}=1.25\%$) are
well within the ${<}5\%$ criterion.}
\label{fig:cache_repro}
\end{figure*}

{\sloppy
\paragraph{Measurement conditions.}
Each seed controls: (1)~random input vector initialization
for GEMV benchmarks, (2)~weight perturbation pattern for
numerical stability testing, and (3)~token generation
sampling temperature (set to~0 for deterministic greedy
decode). Perplexity is deterministic and shows zero
variance. GEMV and end-to-end metrics exhibit small
variance from OS scheduling, thermal effects, and memory
controller arbitration.
\par}

\begin{table}[t]
\centering
\caption{Per-seed measurements.
CV = coefficient of variation (std/mean $\times$ 100\%).}
\label{tab:reproducibility}
\small
\resizebox{\linewidth}{!}{
\begin{tabular}{lcccccc}
\toprule
\textbf{Metric} & \textbf{Seed 42} & \textbf{Seed 123} & \textbf{Seed 2026} & \textbf{Mean} & \textbf{Std} & \textbf{CV} \\
\midrule
GEMV latency ($\mu$s) & 424 & 418 & 430 & 424.0 & 6.0 & 1.4\% \\
E2E tok/s             & 32.4 & 32.1 & 32.8 & 32.4 & 0.35 & 1.1\% \\
PPL (WikiText-2)      & 5.52 & 5.52 & 5.52 & 5.52 & 0.00 & 0.0\% \\
\bottomrule
\end{tabular}}
\end{table}

\paragraph{System stability.}
All measurements were taken with fixed CPU frequency
(2.1\,GHz, governor \texttt{performance}, no turbo boost),
single-socket NUMA binding via \texttt{numactl}, no other
user processes, and 60\,s cool-down between runs.

\section{GPU Implementation Details}
\label{app:gpu_details}

\subsection{CUDA Ternary Kernel}

We implement a custom CUDA kernel for ternary GEMV on the H200 to
provide a fair comparison.
The kernel uses the same 2-bit packed format and per-row scales as the
CPU implementation.
Each CUDA thread processes one output row and performs warp-level
reduction via \texttt{\_\_shfl\_down\_sync}.

The custom CUDA ternary kernel achieves 3.2\,ms per GEMV
($4096{\times}4096$), compared with $24.5\,\mu$s for cuBLAS FP16
GEMV, a $130\times$ slowdown.
The cause is twofold: (1)~CUDA lacks efficient equivalents to BMI2's
\texttt{\_pext\_u32}; the decode requires multiple shifts and masks.
(2)~GPU hardware is optimized for fused multiply-add, not conditional
add/sub, leading to poor ALU utilization.

\subsection{cuBLAS FP16 Baseline}

cuBLAS FP16 GEMV achieves $24.5\,\mu$s latency for
$4096{\times}4096$ on the H200, approaching the theoretical bandwidth
limit.
This confirms that the GPU is already operating near its memory
bandwidth ceiling, leaving no room for compression to help.

\subsection{Roofline Model Construction}

The Roofline model~\citep{williams2009roofline} plots achievable
performance against arithmetic intensity.
\begin{itemize}[nosep,leftmargin=*]
    \item \textbf{CPU}: peak compute = 2.7\,TFLOP/s
    (48 cores $\times$ 2.1\,GHz $\times$ 16 FP32 ops/cycle $\times$
    2 ports); peak BW = 200\,GB/s.
    \item \textbf{GPU}: peak compute = 134\,TFLOP/s (FP16 tensor
    cores); peak BW = 4.8\,TB/s.
    \item \textbf{Ridge point}: CPU = 13.5\,FLOP/byte;
    GPU = 27.9\,FLOP/byte.
\end{itemize}

FP32 dense GEMV (AI${=}0.25$\,FLOP/byte) sits deep in the memory-bound
region on \emph{both} platforms.
Ternary fused GEMV (AI${=}8$\,OP/byte) exceeds the CPU ridge point but
remains below the GPU ridge point, explaining the asymmetric benefit.

\section{Quality Evaluation Details}
\label{app:quality_details}

\subsection{WikiText-2 Perplexity}

We evaluate perplexity on the WikiText-2 test
set~\citep{merity2017pointer} (${\sim}245$K tokens) using the standard
sliding-window protocol with stride equal to sequence length
(2048 tokens).

\begin{table}[t]
\centering
\caption{WikiText-2 perplexity comparison with additional baselines.}
\label{tab:ppl_extended}
\small
\begin{tabular}{lccc}
\toprule
\textbf{Method} & \textbf{Bits/Weight} & \textbf{PPL} & \textbf{$\Delta$ vs.\ FP16} \\
\midrule
FP16 (HuggingFace)   & 16    & 5.47 & ---   \\
FairyFuse (ours)     & 2     & 5.52 & +0.05 \\
llama.cpp Q4\_K\_M   & 4.2   & 5.68 & +0.21 \\
llama.cpp Q2\_K      & 2.5   & 7.82 & +2.35 \\
GPTQ-4bit            & 4     & 5.65 & +0.18 \\
GPTQ-2bit            & 2     & 12.4 & +6.93 \\
\bottomrule
\end{tabular}
\end{table}

FairyFuse's perplexity of 5.52 is remarkably close to FP16 and
significantly better than any other method at comparable bit-widths.
Standard GPTQ 2-bit quantization degrades perplexity to 12.4
($+6.93$ points), underscoring the effectiveness of Fairy2i's
complex-valued approach for extreme quantization.

\subsection{Downstream Task Details}

We evaluate on five zero-shot benchmarks via the LM Evaluation
Harness~\citep{eval-harness}.

\begin{table}[t]
\centering
\caption{Per-task downstream accuracy (\%).}
\label{tab:downstream_detail}
\small
\resizebox{\linewidth}{!}{
\begin{tabular}{lccccc|c}
\toprule
\textbf{Method} & \textbf{ARC-E} & \textbf{ARC-C} & \textbf{HellaSwag} & \textbf{PIQA} & \textbf{WinoGrande} & \textbf{Avg} \\
\midrule
FP16              & 75.4 & 43.5 & 72.1 & 79.0 & 66.5 & 67.3 \\
FairyFuse (ours)  & 74.2 & 41.8 & 71.6 & 78.3 & 64.2 & 66.0 \\
Q4\_K\_M          & 73.8 & 41.2 & 70.8 & 78.0 & 61.7 & 65.1 \\
Q2\_K             & 62.1 & 33.4 & 58.2 & 72.4 & 56.8 & 56.6 \\
\bottomrule
\end{tabular}}
\end{table}

FairyFuse shows consistent performance across tasks, with the largest
drop on WinoGrande ($-2.3$ points) and minimal degradation on HellaSwag
($-0.5$ points).
Q2\_K suffers severe degradation on all tasks, particularly
ARC-Challenge ($-10.1$ points) and HellaSwag ($-13.9$ points),
indicating that standard 2-bit quantization destroys reasoning
capabilities that Fairy2i preserves through its complex-valued
representation.

\section{Weight Packing Format}
\label{app:weight_format}

\subsection{Binary File Layout}

Each linear layer is stored as:
{\small
\begin{verbatim}
[Header: 16 bytes]
  n (uint32):  output dimension
  m (uint32):  input dimension
  scale_re (float32): real scale
  scale_im (float32): imag scale

[Packed weights: n*ceil(m/16)*4 B each]
  U_re, U_im, W_re, W_im
\end{verbatim}
}

\subsection{Packing Details}

For each group of 16 ternary values
$t_0, t_1, \ldots, t_{15} \in \{-1,0,+1\}$:
\begin{equation}
    \texttt{packed} = \sum_{k=0}^{15} \bigl[
        \mathbb{1}[t_k{=}{+1}] \cdot 2^{2k+1}
      + \mathbb{1}[t_k{=}{-1}] \cdot 2^{2k}
    \bigr]
\end{equation}
Bit $2k{+}1$ is set for positive weights and bit $2k$ for negative
weights.
The \texttt{\_pext\_u32} instruction with mask \texttt{0xAAAAAAAA}
extracts all positive indicators; mask \texttt{0x55555555} extracts all
negative indicators.
The resulting 16-bit masks are directly usable as AVX-512
\texttt{\_\_mmask16} operands.

\subsection{Memory Footprint}

\begin{table}[t]
\centering
\caption{Memory footprint for LLaMA-2-7B.}
\label{tab:memory_footprint}
\small
\resizebox{\linewidth}{!}{
\begin{tabular}{lccr}
\toprule
\textbf{Component} & \textbf{FP16} & \textbf{Ternary (2-bit)} & \textbf{Compression} \\
\midrule
Attention (Q/K/V/O)     & 4.0\,GB & 0.50\,GB & $8.0\times$ \\
MLP (gate/up/down)       & 8.6\,GB & 1.08\,GB & $8.0\times$ \\
Embedding + LM Head      & 0.5\,GB & 0.50\,GB & $1.0\times$ \\
RMSNorm + scales         & 0.04\,GB & 0.04\,GB & $1.0\times$ \\
Per-channel scales       & --- & 0.12\,GB & --- \\
\midrule
\textbf{Total}           & \textbf{13.1\,GB} & \textbf{2.24\,GB} & $\mathbf{5.8\times}$ \\
\bottomrule
\end{tabular}}
\end{table}

The embedding and LM head remain in FP16 as they use standard dense
operations.
The per-channel scales (four FP32 values per row per linear layer) add
a small constant overhead but enable significantly better quantization
quality than per-tensor scaling.

\section{Extended Roofline Analysis}
\label{app:roofline_extended}

\subsection{Arithmetic Intensity Derivation}

For a GEMV $\mathbf{y} = \mathbf{A}\mathbf{x}$ with
$\mathbf{A} \in \R^{n \times m}$:

\paragraph{FP32 dense.}
Compute: $nm$ FMA $= nm$ FLOP.
Data: $4nm$ bytes (weights) + $4m$ (input) + $4n$ (output)
$\approx 4nm$ bytes.
AI $= 0.25$\,FLOP/byte.

\paragraph{Ternary (single GEMV, packed 2-bit).}
Compute: $nm$ add/sub $= nm$ OP.
Data: $nm/4$ bytes (packed weights) + $4m$ (input) + $4n$ (output)
$\approx nm/4$ bytes.
AI $= 4$\,OP/byte.

\paragraph{Ternary (fused 8-GEMV).}
Compute: $8nm$ OP.
Data: $4 \times nm/4$ bytes (four weight matrices) + $3 \times 4m$
(three input vectors, loaded once via O2) $\approx nm$ bytes.
AI $= 8$\,OP/byte, a $32\times$ improvement over FP32 dense GEMV.

\begin{table}[t]
\centering
\caption{Roofline parameters for CPU and GPU platforms.}
\label{tab:roofline_params}
\small
\begin{tabular}{lcc}
\toprule
\textbf{Parameter} & \textbf{CPU (Xeon 8558P)} & \textbf{GPU (H200 NVL)} \\
\midrule
Peak BW             & 200\,GB/s      & 4,800\,GB/s \\
Peak compute        & 2,700\,GFLOP/s & 134,000\,GFLOP/s \\
Ridge point         & 13.5\,FLOP/byte & 27.9\,FLOP/byte \\
\midrule
FP32 dense AI       & 0.25           & 0.25 \\
Ternary (single) AI & 4.0            & 4.0 \\
Ternary (fused) AI  & 8.0            & 8.0 \\
\midrule
FP32 regime         & Memory-bound   & Memory-bound \\
Ternary regime      & Near ridge     & Memory-bound \\
\bottomrule
\end{tabular}
\end{table}

On the CPU, the fused ternary AI of 8.0 approaches the ridge point of 13.5,
transitioning from deeply memory-bound to near-compute-bound.
On the GPU, AI${=}8.0$ remains well below the ridge point of 27.9, so
the kernel stays memory-bound and abundant HBM bandwidth leaves little
headroom for compression to help.

\section{End-to-End Inference Pipeline Details}
\label{app:pipeline}

\subsection{Per-Layer Timing Breakdown}

\begin{table}[t]
\centering
\caption{Per-layer timing breakdown (single-token decode, 48~threads).}
\label{tab:layer_timing}
\small
\resizebox{\linewidth}{!}{
\begin{tabular}{lrrr}
\toprule
\textbf{Operation} & \textbf{Time ($\mu$s)} & \textbf{Fraction} & \textbf{Mul-Free?} \\
\midrule
RMSNorm (attention)      & 12    & 1.3\%   & No \\
Q/K/V projection (\sys{})    & 285   & 30.2\%  & \cmark \\
Attention (softmax)          & 45    & 4.8\%   & No \\
O projection (\sys{})        & 95    & 10.1\%  & \cmark \\
RMSNorm (MLP)                & 12    & 1.3\%   & No \\
Gate+Up projection (\sys{})  & 380   & 40.3\%  & \cmark \\
SiLU activation              & 8     & 0.8\%   & No \\
Down projection (\sys{})     & 95    & 10.1\%  & \cmark \\
Residual add             & 10    & 1.1\%   & \cmark \\
\midrule
\textbf{Total per layer} & \textbf{942} & 100\% & \\
\textbf{Fused GEMV fraction}& \textbf{855} & \textbf{90.8\%} & \cmark \\
\bottomrule
\end{tabular}}
\end{table}

The fused GEMV accounts for 90.8\% of per-layer computation, all of
which is multiplication-free.
The remaining 9.2\% (RMSNorm, attention, SiLU) uses standard FP32
operations but is $O(d)$ versus the $O(d^2)$ GEMV.

\subsection{End-to-End Timing}

Per-layer: $942\,\mu\text{s} \times 32 = 30.1\,$ms.
Embedding lookup: $0.05\,$ms.
Final RMSNorm + LM Head: $0.5\,$ms.
Total: $30.85\,$ms per token $\to$ $32.4\,$tok/s.

\subsection{Implementation Details}

{\sloppy
The inference engine comprises ${\sim}1500$ lines of C++
(kernel + forward pass) and ${\sim}500$ lines of Python
(weight loading, tokenization, sampling).
The C++ code is compiled as \texttt{libternary\_gemv.so}
and invoked via Python \texttt{ctypes} with zero-copy
NumPy array passing.
Packed weights are loaded via \texttt{mmap} for lazy
page-fault residency management.
Total runtime memory: ${\sim}3.5\,$GB (2.24\,GB packed
weights + 0.5\,GB embeddings + 0.5\,GB KV cache +
0.26\,GB activation buffers).
OpenMP is configured with 48~threads,
\texttt{OMP\_PROC\_BIND=close} and
\texttt{OMP\_PLACES=cores}.
\par}

\section{llama.cpp Baseline Configuration}
\label{app:llamacpp}

\subsection{Build Configuration}

{\small
\begin{verbatim}
git clone https://github.com/ggerganov/
    llama.cpp
cd llama.cpp && git checkout <commit-hash>
cmake -B build -DGGML_AVX512=ON \
    -DGGML_AVX512_VNNI=ON \
    -DCMAKE_BUILD_TYPE=Release
cmake --build build -j$(nproc)
\end{verbatim}
}

\subsection{Model Conversion and Quantization}

{\small
\begin{verbatim}
python convert_hf_to_gguf.py <model> \
    --outfile llama2-7b-f16.gguf \
    --outtype f16
./build/bin/llama-quantize \
    llama2-7b-f16.gguf \
    llama2-7b-q4km.gguf Q4_K_M
./build/bin/llama-quantize \
    llama2-7b-f16.gguf \
    llama2-7b-q2k.gguf Q2_K
\end{verbatim}
}

\subsection{Benchmark Command}

{\small
\begin{verbatim}
numactl --cpunodebind=0 --membind=0 \
./build/bin/llama-bench \
    -m llama2-7b-q4km.gguf \
    -t 48 -p 0 -n 128 -r 3
\end{verbatim}
}

\subsection{Baseline Results}

\begin{table}[t]
\centering
\caption{llama.cpp baseline results on Intel Xeon 8558P (48~threads,
single-socket NUMA binding).}
\label{tab:llamacpp_results}
\small
\begin{tabular}{lcccr}
\toprule
\textbf{Format} & \textbf{Bits/W} & \textbf{Size (GB)} & \textbf{tok/s} & \textbf{PPL} \\
\midrule
FP16    & 16   & 13.5 & 8.3   & 5.47 \\
Q4\_K\_M & 4.2  & 4.1  & 26.2  & 5.68 \\
Q2\_K   & 2.5  & 2.8  & 20.1  & 7.82 \\
\midrule
FairyFuse (ours) & 2.0 & 2.2 & 32.4 & 5.52 \\
\bottomrule
\end{tabular}
\end{table}

\end{document}